\documentclass[preprint,12pt]{elsarticle}

\usepackage{amssymb}

\usepackage{amsmath}
\usepackage{array}
\usepackage{rotating}
\usepackage{tikz}
\usepackage[utf8]{inputenc} 
\usepackage[T1]{fontenc}    
\usepackage{lmodern}
\usepackage[hidelinks]{hyperref}       
\usepackage{url}            
\usepackage{booktabs}       
\usepackage{amsfonts}       
\usepackage{nicefrac}       
\usepackage{microtype}      
\usepackage{tabularx}
\usepackage{tcolorbox}
\usepackage{subcaption}
\tcbuselibrary{skins}
\usetikzlibrary{positioning}
\usetikzlibrary{backgrounds}
\usepackage{adjustbox}

\usepackage{colortbl}
\usepackage{dsfont}
\usepackage{rotating}
\usepackage{amsthm}
\usepackage{amsmath}       
\usepackage{ifthen}

\usepackage{grffile}
\usepackage{enumitem}
\usepackage{wrapfig}
\usepackage{floatrow}

\newcommand{\defeq}{\mathrel{\overset{\makebox[0pt]{\mbox{\normalfont\tiny\sffamily def}}}{=}}}

\definecolor{babyblue}{rgb}{0.54, 0.81, 0.94}

\newcommand{\E}{\mathbb{E}}

\newcommand{\Actions}{\mathcal{A}}
\newcommand{\States}{\mathcal{S}}

\newcommand{\repfunc}{\Phi_{\theta_R}}
\newcommand{\repfunctarget}{\Phi_{\hat{\theta}_R}}
\newcommand{\auxfunc}{F_{\theta_A}}
\newcommand{\auxfunctarget}{F_{\hat{\theta}_A}}

\journal{ Journal of Artificial Intelligence }

\begin{document}

\begin{frontmatter}



\title{Investigating the Properties of
	Neural Network Representations in Reinforcement Learning}


\author[label1]{Han~Wang} 
\author[label1]{Erfan~Miahi} 
\author[label1]{Martha~White} 
\author[label1,label2]{Marlos~C.~Machado} 
\author[label2]{Zaheer~Abbas}
\author[label3]{Raksha~Kumaraswamy}
\author[label1]{Vincent~Liu}
\author[label1]{Adam~White}

\affiliation[label1]{organization={Department of Computing Science, University of Alberta; \\Alberta Machine Intelligence Institute}, country={Canada}}
\affiliation[label2]{organization={Deepmind}, country={Canada}}
\affiliation[label3]{organization={Sony AI (Work done at University of Alberta)}, country={Canada}}

\begin{abstract}
 In this paper we investigate the properties of representations learned by deep reinforcement learning systems. Much of the early work on representations for reinforcement learning focused on designing fixed-basis architectures to achieve properties thought to be desirable, such as orthogonality and sparsity. In contrast, the idea behind deep reinforcement learning methods is that the agent designer should not encode representational properties, but rather that the data stream should determine the properties of the representation---good representations emerge under appropriate training schemes. In this paper we bring these two perspectives together, empirically investigating the properties of representations that support transfer in reinforcement learning. 
We introduce and measure six representational properties over more than 25 thousand agent-task settings. We consider Deep Q-learning agents with different auxiliary losses in a pixel-based navigation environment, with source and transfer tasks corresponding to different goal locations.  We develop a method to better understand \emph{why} some representations work better for transfer, through a systematic approach varying task similarity and measuring and correlating representation properties with transfer performance. We demonstrate the generality of the methodology by investigating representations learned by a Rainbow agent that successfully transfer across games modes in Atari 2600.

\end{abstract}



\end{frontmatter}


\section{Good Representations for RL}
\label{sec:introduction}
In reinforcement learning an agent interacts with their environment, receiving observations and taking actions based on those observations, with the goal of maximizing the sum of a special numerical signal, the reward. 
The function that converts these observations
is known as \emph{representation}, its elements are known as \emph{features}, and the process of learning such function is known as \emph{representation learning}. Ultimately, many other subproblems depend on the agent's representation. Bad representations hinder predictions and diminish the effectiveness of planning and learning algorithms~\cite{talvitie2017self,holland2018effect,wan2019planning,du2019isa}. Good representations can lead to better sample efficiency~\cite{russo2013eluder,lattimore2020learning}. Therefore, the key question motivating this research is: \emph{what are good representations and how can the agent find them?}

Good representations are classically defined in service to other tasks. Often, good representations are said to be those that improve agents in some dimension, such as learning efficiency \cite{watter2015embed,jaderberg2016reinforcement}, performance in unseen tasks \cite{rusu2017sim,finn2017model,barreto2017successor,higgins2017darla,peng2018sim,zhang2018dissection,nair2018visual,tyo2020transferable}, the accuracy of learned models for planning \cite{silver2017predictron,gupta2017cognitive,srinivas2018universal,kurutach2018learning,franccois2019combined,yang2020plan2vec}, and in the agent's ability to represent the world the way humans do \cite{finn2016deep,higgins2018towards}. 
In this context, there are two main approaches for obtaining good representations: using a fixed, expert-designed transformations of the agent's observations, or learning such transformations from data. 

Fixed transformations of the agent's observations, which lead to fixed-basis architectures, have been extensively explored in reinforcement learning. They allow us to enforce specific properties that  are thought to be beneficial. For example, many approaches either use or search for orthogonal or decorrelated features, such as orthogonal matching pursuit~\cite{painter2012greedy}, Bellman-error basis functions \cite{parr2007analyzing}, Fourier basis \cite{konidaris2011value}, tile coding \cite{sutton1996generalization}, and proto-value functions \cite{mahadevan2007proto,kompella2012incremental}. 
Prototypical input matching methods have been explored, as in kernel methods, radial basis functions \cite{sutton1998reinforcement}, cascade correlation networks \cite{fahlman1990cascade}, and Kanerva coding \cite{kanerva1988sparse}. 
Some of these approaches produce high-dimensional, sparse representations, which are more likely to be orthogonal. Moreover, by activating only a small subset of features, a sparse representation reduces computation and increases scalability, such as in tile coding \cite{sutton1996generalization} and in sparse distributed memories \cite{ratitch2004sparse}.
However, fixed-basis architectures are not adaptive and are difficult to scale to high-dimensional inputs.

Recent developments in representation learning for reinforcement learning explore a different perspective: we should avoid optimizing specific properties\footnote{There is, of course, work in reinforcement learning exploring how to encode specific properties on the network (e.g., sparse activations \cite{liu2019utility, pan2021fuzzy}, disentangled features \cite{higgins2017darla}, and orthogonality constraints \cite{wu2018laplacian,franccois2019combined}).} and instead use gradient descent to let the training data dictate the properties of the representation. 
This is achieved with specific training regimes, including multi-task (parallel) training \cite{french1999catastrophic,caruana1997multitask,thrun1996learning}, auxiliary losses~\cite{jaderberg2016reinforcement,bellemare2019geometric}, and training on a distribution of problems (\`a la meta-learning)~\cite{finn2017model,silver2017predictron,oh2017value,schrittwieser2020mastering,javed2019meta}. The underlying idea is that good representations will emerge if the problem setting is complex enough, where goodness is often measured by success on some held-out test task.  

There are many different ways to evaluate and understand these emergent representations.
Recent work has explored this question in roughly two ways: what do good representations look like, and what capabilities do good and bad representations allow. The most common approach is to visualize the learned representations \cite{mnih2015human,watter2015embed,gupta2017cognitive,rupprecht2020visualizing,zahavy2016graying,finn2016deep,greydanus2018visualizing,such2018atari,kurutach2018learning,srinivas2018universal,dai2019analysing,franccois2019combined,bellemare2019geometric}. This approach has been used, for example, to provide evidence for the emergence of abstraction and compositionality in supervised learning~\cite{yosinski2015understanding,nguyen2017plug}. However, in reinforcement learning, the impact of delayed consequences and temporally correlated data make it difficult to import these analysis techniques from other fields, and recent work highlighted how popular approaches like saliency maps may not always be appropriate~\cite{atrey2019exploratory}.

This paper explores the properties of representations learned by deep reinforcement learning systems: specifically (1) DQN~\cite{mnih2015human}---Q-learning with neural network function approximation---combined with different auxiliary tasks, and (2) Rainbow ~\cite{hessel2018rainbow}. 
We investigate properties grouped in three categories: \emph{capacity} (complexity reduction, dynamics awareness, and diversity), \emph{redundancy} (orthogonality and sparsity), and \emph{robustness} (non-interference). 
Our property set consists of both a subset of properties discussed in the literature, and properties newly introduced in this paper. We focus on the fully observable setting; our goal is to understand how the representation transforms the current input. Other properties and experiments would be suitable to understand (recurrent) representations that summarize histories for the partially observable setting, but this is left to future work. \looseness=-1 


We focus on a {\em representation transfer} setting: a training phase to learn a representation in one {\em source task}, followed by testing in a related {\em transfer task} where the agent learns using that fixed, pre-learned representation. Our primary goal is to identify representations that are useful for future learning.
Ideally, it
should be useful for future learning in the same task. Additionally, the pre-learned representation should also be useful later when learning about other related tasks. Such a representation should enable faster learning, compared with re-learning everything from scratch, even if we prevent the agent from changing the representation after pre-training---perhaps the clearest and sternest evaluation of the usefulness of prior learning. If fine-tuning or transfer of both the representation and the value function are required to outperform learning from scratch, then one might fairly wonder if any representation is learned at all.


Representation transfer, as defined above, relies on a set of related tasks to evaluate future learning.
We start with a simple image-based maze environment, that is computationally efficient and so allows exhaustive experiments, as well as provides a natural notion of task similarity (goal location). We then test across different game modes in Atari 2600 games~\cite{bellemare2013the}. Recent work has demonstrated that transfer between games modes is possible with a variant of the Rainbow agent, \cite{rusu2022probing}, however transfer is clearly nontrivial as prior investigations with DQN reported failure \cite{farebrother2018generalization}.

Our first study in the image-based maze environment uses nine auxiliary tasks, resulting in 150 representations for 173 target tasks. 
We investigated two activations: the widely used ReLU which produces a relatively compact, dense representation, and a new activation function called FTA \cite{pan2021fuzzy} that produces high-dimensional, sparse representations. 
The key insights are as follows.
\begin{enumerate}[leftmargin=*]
	\item Auxiliary tasks can facilitate emergence of representations effective for transfer, however with ReLU networks many auxiliary tasks do not outperform learning from scratch (do not transfer).
	\item Using sparse activations (FTA) was a significant factor in improving transfer. The FTA-based representations transferred consistently, with or without auxiliary tasks. 
	\item ReLU-based representations transfered well to very similar tasks (better than FTA), but significantly worse than FTA to less similar tasks in our setup. 
	\item Transfer was not possible with linear value functions: performance was significantly better when the representation was inputted into a nonlinear value function. 
	\item The representations that transferred best had high levels of complexity reduction, medium-high levels of dynamics awareness and diversity, and medium levels of orthogonality and sparsity. 
\end{enumerate}
A key contribution of this work is providing a systematic approach to investigate representations and their properties. The empirical design took many iterations, including 1) developing the transfer setup and a novel way of ranking task similarity using successor features so that we could systematically vary the level of difficulty in transfer, 2) developing the set of properties to measure, 3) appropriately sweeping hyperparameters to obtain reasonably performing agents but still avoiding over-tuning, and 4) providing several mechanisms to aggregate and visualize the mountain of data produced across representations. For example, initial results had little consistency because the agents themselves were not effectively trained; it turns out analyzing poorly performing agents results in unclear conclusions. 

Using these insights, we applied our methodology to understand representation transfer across different Atari 2600 game modes~\cite{bellemare2013the,machado2018revisiting}. We trained a Rainbow agent on the default game mode, and showed that the learned representation facilitated transfer to other modes in three different games. We found similar outcomes in terms of the properties and transfer performance, which is both somewhat surprising and an indicator that the properties  and methodology we propose here are meaningful. More specifically, we similarly found that the Rainbow representation had 
\begin{enumerate}[leftmargin=*]
	\item high complexity reduction---increasing this level significantly from its random initialization
	\item high orthogonality and sparsity, even more so than the representations in the maze environment
	\item medium diversity 
\end{enumerate}


The results in this work complement the growing literature on representation transfer in reinforcement learning by providing a quantitative approach to understand the learned representations. We start the paper by more explicitly defining representations (Section 2) and what it means to be a good representation (Section 3). We then explain the auxiliary losses (Section 4) that result in the many different representations we analyze in the results. We then explain the experimental setup (Section 5) and provide our first key result showing transfer performance, both successful and failed, for these many representations (Section 6). We then introduce the properties (Section 7) and then analyze these properties (Section 8) showing correlations to transfer performance (using 150 representations), the evolution of property values over time and finally a contrast between final property values for three representations that provided the bulleted conclusions above. We conclude with results in Atari (Section 9). \looseness=-1

\section{Problem Formulation and Notation}

We formalize the agent's interaction as a finite Markov Decision Process (MDP) with a finite state space $\States$, finite action space $\Actions$, transition function $P: \States \times \Actions \times \States \rightarrow [0, 1]$, and bounded reward function $R: \States \times \Actions \times \States \rightarrow \mathbb{R}$. On each time step, $t=1,2,...$, the agent takes action $A_t$ in state $S_t$ and the environment transitions to state $S_{t+1} \sim P(\cdot | S_t, A_t)$ and emits a reward $R_{t+1}$. 
The agent's objective is to find a policy, $\pi: \States \times \Actions \rightarrow [0,1]$ that maximizes the expected discounted sum of future rewards, the \emph{return}, 
$G_t \doteq R_{t+1} + \gamma_{t+1} G_{t+1}$,
where $\gamma_{t+1} \in [0,1]$ denotes a discount that depends on the transition $(S_t, A_t, S_{t+1})$~\cite{white2017unifying}. In episodic problems, which we study here, the discount might be 1 during the episode, and it becomes zero when $S_t, A_t$ lead to termination.

For much of this work we study the representations learned by DQN~\cite{mnih2015human}, a widely used value-based method in deep reinforcement learning. (In Section \ref{sec_atari}, we study Rainbow~\cite{hessel2018rainbow}, and leave the description to then.) 
The approximate value function is parameterized by a set of weights $\theta$: $Q_\theta(s,a) \approx q_\pi(s,a)$. The word \emph{deep}, in deep reinforcement learning, stems from the use of neural networks to approximate $q_\pi$. In particular, DQN iteratively updates its action-value estimates by training the parameters of a neural network, $\theta$, with stochastic gradient descent: $\Delta\theta \propto \big(R_{t+1} + \gamma_{t+1}\max \bar{Q}(S_{t+1}, a) - Q_\theta (S_t, A_t)\big)\nabla_\theta Q_\theta (S_t, A_t)$. The target network, $\bar{Q}$, is not updated on every step, but only periodically set equal to the current $Q_\theta$.  
Actions are selected according to an $\epsilon$-greedy policy, where $A_t\doteq \arg\max_{a \in \mathcal{A}}Q_\theta(S_t, a)$ with probability $1-\epsilon$ or a random action with probability $\epsilon$. As is typical, we use mini-batch updates from an experience replay buffer~\cite{lin92self}. \looseness=-1

\begin{figure}[t]
	\centering
	\vspace{-0.5cm}
	\begin{tikzpicture}
	\node (img)  {\includegraphics[width=0.85\linewidth]{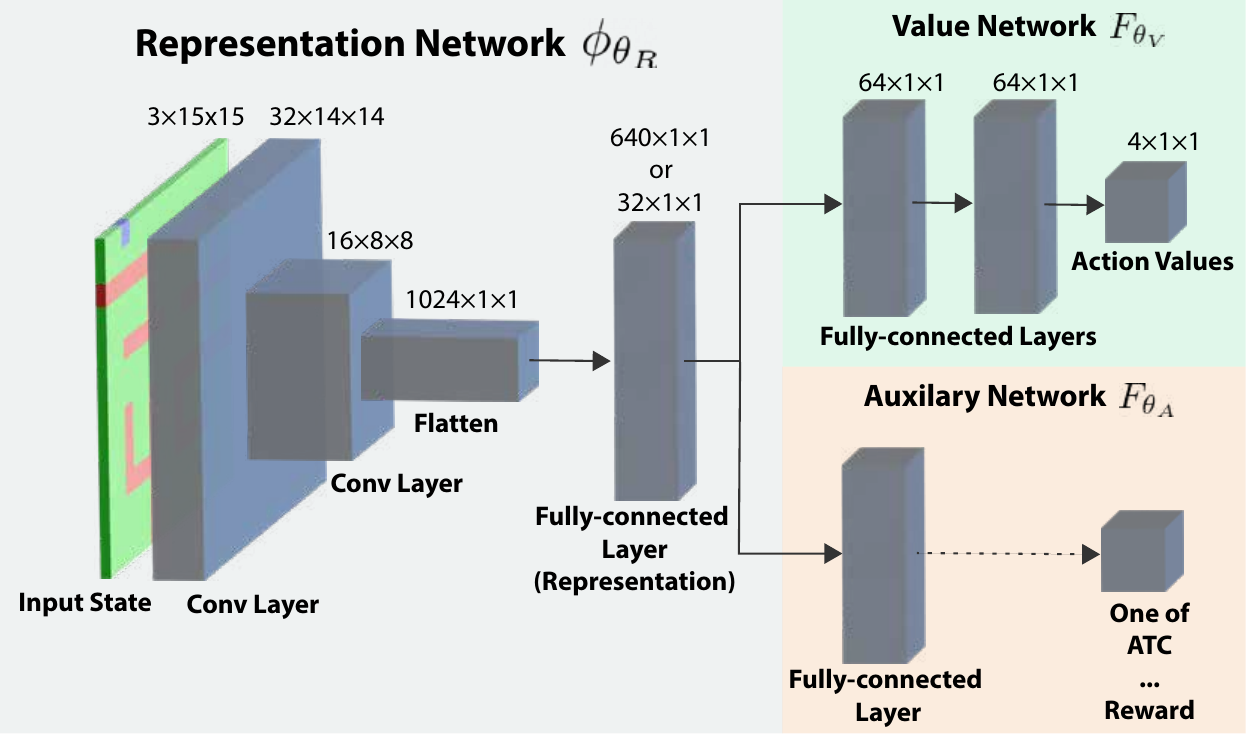}};
	\end{tikzpicture}
	\caption{We experiment with agents using this network architecture, with different auxiliary losses. The representation network, $\repfunc$, learns a mapping from input-state $s_t$ to the agent-state (representation of $s_t$). The representation network is learned to improve two objectives: performance on a main task and on an auxiliary task. The diagram depicts the auxiliary tasks we use in this work, described in Section~\ref{sec:auxlosses}. Our agents only use one auxiliary task at a time.
	}
	\label{fg:overview}
	\vspace{-0.4cm}
\end{figure}

We use image inputs, a convolutional network and a fully connected layer with a lower dimension $d$. We consider two activations on this fully connected layer: rectified linear units (ReLU) \cite{nair2010rectified} and the fuzzy tiling activation (FTA) \cite{pan2021fuzzy}. FTA is a one-to-many activation that leads to a larger number of units in the representation layer: $k \times d$ instead of $d$, but with only a small number of active features at once.
This allows us to investigate more compact, lower-dimensional representations produced by the ReLU and higher-dimensional, sparse representations produced by FTA. We call these representations the \emph{representation layer}.  

We make extensive use of auxiliary tasks~\cite{jaderberg2016reinforcement} to both induce better representations and to study them. 
Auxiliary tasks are additional prediction tasks given to agents to incentivize the network to learn about properties of the environment which are, in principle, not directly related to reward maximization. Examples include predicting pixel changes~\cite{jaderberg2016reinforcement} and the next state given an action~\cite{oh2015action}. These tasks are posed as additional loss functions and the agent is tasked to balance between them and~return~maximization.

We use a unified architecture to explore representations induced by a variety of different auxiliary tasks, as shown in Figure \ref{fg:overview}. 
The first layers, parameterized by $\theta_R$, produce the representation $\phi_t=\repfunc(s_t)$. 
The last layers, which are parameterized with $\theta_V$, use the representation to estimate the action-values. 
Auxiliary tasks are encoded with additional layers and separate heads (with parameters $\theta_A$), further impacting the updates to $\theta_R$ via gradient descent: $\repfunc(s)$ must be adjusted to be useful for both estimating action-values and reducing the auxiliary losses.

Given this setup, a natural (and basic) question is: \emph{what is the representation in a deep RL agent?} The answer relies on the role of the representation, which is primarily to promote \emph{future learning}. We learn a transformation on the inputs---features---to facilitate downstream learning. 
The features should 1) be \emph{reusable} or generally useful for multiple predictions, 2) improve \emph{sample efficiency} for the given online algorithm (e.g., SGD), and 3) be \emph{computationally efficient}.\footnote{Another criteria that has been considered for partially observable settings, especially where the true state is low-dimensional and compact, is how well the representation captures the true state, with related ideas about finding a compact set of disentangled features or causal features; see \cite{lesort2018state} for a nice overview. Such representations could facilitate all three of these goals---reusability, sample efficiency and computational efficiency---but may not be necessary to achieve them.} 

For example, an agent may want to pickup objects in a room, with new objects continually added. The features could describe the objects, so that new objects \emph{reuse} previously learned concepts (e.g., red, cup) that are a succinct (efficient) description of the object. If the agent uses online updating, from temporally correlated samples, we may prefer sparsely activated features that only change a subset of the weights, to reduce interference and promote \emph{sample efficiency}. 

Of course, this is only a hypothetical example; we do not truly know the semantics of what is learned, nor what features would improve learning. In this paper, we attempt to systematically measure representation properties and relationships to transfer performance (future learning), to gain more insight into the representations of deep RL agents.

\section{Good representations for Transfer}\label{sec:method}
We aim to understand the properties of representations that emerge in deep reinforcement learning, but it is critical to do so for both good and bad representations; which, again, begs the question: \emph{what is a good representation}? We use a simple definition: a good representation is one that transfers. More precisely, if features learned during an initial learning phase allow for faster learning on future data, then those features transfer. Good representations that reliably achieve good transfer may exhibit properties and attributes different than representations that result in poor or negative transfer. 

We seek to empirically relate representation properties and performance, which requires an environment where transfer is possible.    
We investigate how the complexity of the value function interacts with transfer in a simple pixel-based navigation environment with obstacles.  This environment, depicted in Figure \ref{fg:environments}, can be readily used to generate numerous related tasks. The agent must learn to navigate to a given goal state in as few steps as possible.
The problem is episodic, with $\gamma = 0.99$, a reward of $+1$ when reaching the goal and 0 otherwise. The input state consists of an RGB input of a $15\times 15$ grid (size $15\times 15\times 3$), encoding the agent's current location (but not the goal). The actions correspond to the four cardinal directions, and transitions the agent deterministically by one pixel, or not at all if the action is into a wall. To simplify exploration, the agent starts in a uniform random state and episodes are cut-off at $100$ steps; the agent is then teleported to a new random state and this transition is discarded.

\definecolor{dark-green}{RGB}{0,100,0}
\definecolor{dark-gray}{RGB}{100, 118, 135}
\definecolor{dark-purple}{RGB}{128,0,128}

\begin{figure}
	\centering
	\begin{subfigure}{.3\columnwidth}
		\centering
		\includegraphics[width=0.9\linewidth]{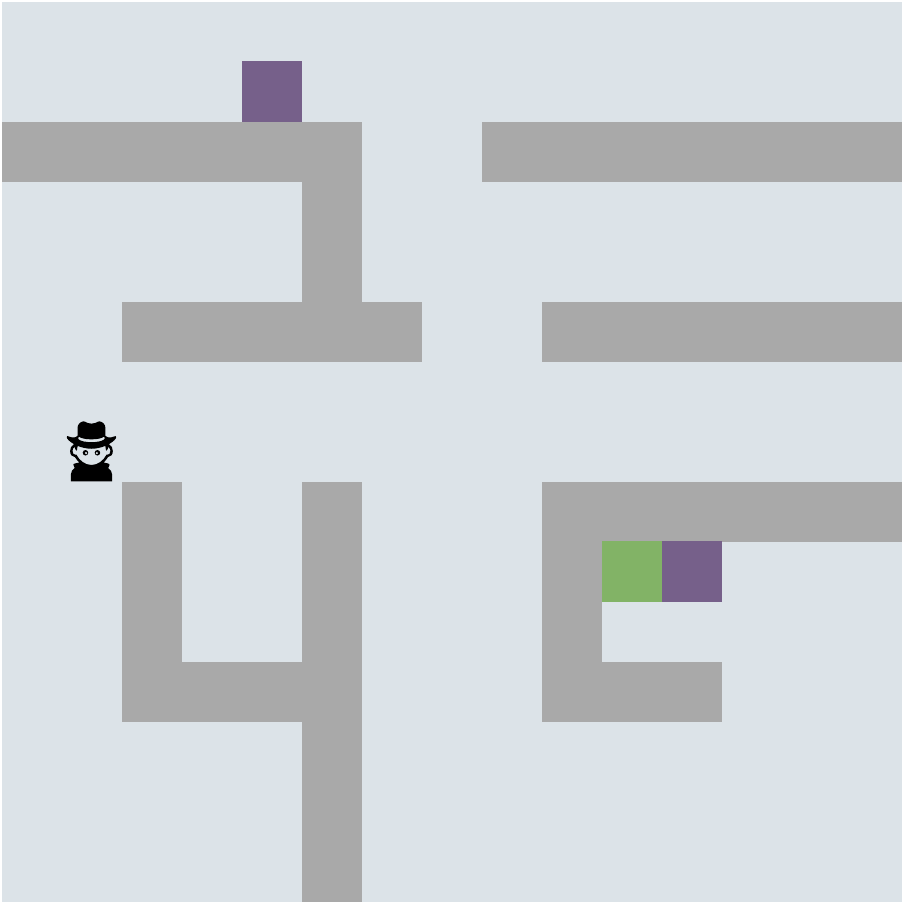}
		\caption{Maze problem}
		\label{fg:environments}
	\end{subfigure}%
	~
	\begin{subfigure}{.55\columnwidth}
		\centering
		\vspace{-0.2cm}
		\includegraphics[width=0.99\linewidth]{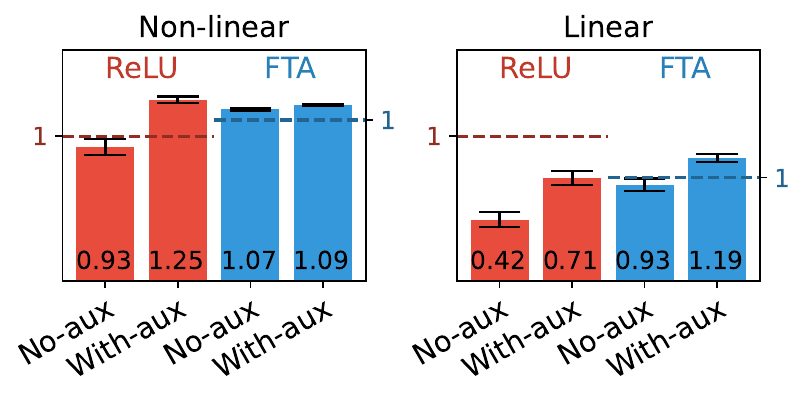}
		\vspace{-0.3cm}
		\caption{Relative Performance}
		\label{fg:result1}
	\end{subfigure}
	\caption{Representation transfer is possible in the Maze navigation task and auxiliary tasks improve transfer when using non-linear function approximation. (a) The position of the {\color{dark-gray} walls (dark grey)}, the {\color{dark-green} goal (green)} in the training, and two {\color{dark-purple} transfer tasks (purple)} are shown. (b) Performance relative to the baseline (from scratch) agent: above $1$ represents an improvement and below $1$ denotes negative transfer. Error bars show a 95\% confidence interval. Performance is reported for the best auxiliary task for each activation: VirtualVF5 for ReLU-based representations, and SF for FTA, explained in Section \ref{sec:auxlosses}. The dotted lines labeled with a $1$ are at different locations, to indicate the relative performance between FTA and ReLU from scratch. In non-linear, FTA from scratch was better; in linear, ReLU from scratch was better.
	}
\end{figure}

We use different navigation tasks (different goal locations) to define training and transfer in our environment. The agent is first trained to go to the goal at a specific location (e.g., $[9, 9]$, depicted in Figure \ref{fg:environments}). Next, we create a new agent with the parameters of the network up to the representation layer copied over from the training agent and we freeze them to prevent further adaption---a {transfer agent}. In this transfer phase of the experiment, the new agent is trained to navigate to a nearby but different goal location. We compare this to a baseline agent where all parameters of the network are not pre-trained but instead randomly initialized and updated during the transfer phase. This baseline agent learns the task from {\em scratch} because it does not benefit in any way from prior learning. If the transfer agent learns more quickly than the baseline scratch agent, then we say the representation learned in the first phase of the experiment {\em facilitates transfer}. Because of the multiple possible goal locations, there are many possible training and transfer tasks in this environment. 

At this point we have the major ideas in place to present a simple but foundational result: \emph{representation transfer is possible in our navigation task and auxiliary tasks improve transfer}. Figure \ref{fg:result1} summarizes this result upfront; we give details later about the experiment as we proceed to more complex and nuanced results. Representations pre-trained on the training task significantly outperformed representations learned from scratch on the transfer task. In addition, we found auxiliary tasks were important for transfer, at least for ReLU. Pre-trained representations using ReLU exhibited negative transfer, whereas ReLU-based representations combined with well-designed auxiliary tasks did transfer. 

These results, though, produced some surprises. First, even in this simple environment, transfer with ReLU and auxiliary tasks was difficult. This reaffirms some of the previous anecdotal and documented \cite{farebrother2018generalization,tyo2020transferable} issues with transferring representations. In fact, switching to a different (sparse) activation, FTA, had a bigger impact than any auxiliary task. This suggest some issues with the representations learned under ReLU. 

Second, \emph{we were unable to obtain successful transfer with linear value functions}. This outcome was not for a lack of effort.
Figure \ref{fg:result1} strongly suggests that, in our navigation environment, {non-linear value functions significantly improve transfer}, even in a relatively simple environment. We see that when the value function was linear in the features, neither a pre-trained representation (labelled ReLU and FTA) nor any other representations trained with auxiliary tasks improved over training a fresh representation from scratch. This suggests representations may emerge in earlier layers of the network, and that it may be more feasible to learn re-usable features when they can be nonlinearly combined, even if only with a simple shallow network. We highlight this important point here; for the remainder of this paper, we restrict our scope only to non-linear value functions. 

We presented these overall results upfront, before diving deeper and understanding \emph{why} we see these outcomes. In the next sections, we describe the auxiliary tasks we used to find good representations; then how we evaluate the representations for different transfer tasks; and finally give insights into failures in transfer under ReLU due to transfer task similarity. Later, we dive even deeper, measuring and correlating representation properties to transfer performance.

\section{The Auxiliary Losses}\label{sec:auxlosses}

We focus on the representations that emerge under different auxiliary losses for two reasons: 1) 
Adding or changing auxiliary tasks does not change the functional capacity and size of the learned representations; the consistency on the capacity makes comparisons more interpretable. 2)~The role and impact of auxiliary losses in reinforcement learning remains poorly understood, and constitutes an important area of study. 
We introduce the general idea behind each auxiliary task in text, while their formal definition is presented in Table~\ref{table:aux-loss} in Appendix \ref{app_aux}.

\textbf{Input Reconstruction (IR):}
This auxiliary task tries to reconstruct the network's input, as in an autoencoder.
This extraction is achieved by using a bottleneck layer: a low-dimensional layer that forces only the most important information to be retained and the remainder, including the noise, to be discarded. We include this auxiliary task as a classic and simple choice.

\textbf{Next Agent State Prediction (NAS):} Another common choice is to predict the next agent state \cite{jaderberg2016reinforcement,pathak2017curiosity,chung2019two,machado2018eigenoption,franccois2019combined,yang2020plan2vec,srinivas2018universal,oh2017value,schrittwieser2020mastering}. This loss encourages the representation to capture the transition dynamics. The agent predicts $\phi_{t+1}$ using $\phi_t$ and action $a_t$. Predicting the next agent state might give vacuous solutions when it is the only training signal; jointly training with the main task, however, prevents this from happening. The combination of this auxiliary loss with the main task encourages the representation to both be useful for action-value estimation, as well as capable of anticipating features on the next step. Several papers have highlighted that the ability to predict the next state is related to the ability to predict action-values \cite{bellemare2019geometric,parr2008linearmodels,szepesvari2010algorithms}. 

\textbf{Successor Feature Prediction (SF):} NAS can be taken one step further, with the target including not just the next agent-state but many future agent states \cite{machado2020count}. \emph{Successor features}~(SFs) provide just such a target. SFs are defined with respect to a particular policy $\pi$ as $\psi^\pi_t = \E \big[\sum_{i=0}^\infty \gamma^{i} \phi_{t+i}\big]$. They have been used in the transfer setting because they can be used to quickly infer value estimates for new reward functions that are a linear function $\phi_t$ \cite{barreto2017successor}. 
In the tabular case, SFs correspond to successor representations~\cite{dayan1993improving}, which have an equivalence to proto-value functions \cite{machado2018eigenoption}.
We opt to use the greedy policy according to the action values for the main task, which means the SFs are tracking a changing policy.

\textbf{Reward Prediction (Reward):} Another auxiliary task we consider is predicting the immediate reward in the future based on the current state and action \cite{jaderberg2016reinforcement}. The prediction requires the agent to encode the reward information that it can obtain in a short term in the representation function. 

\textbf{Expert Target Prediction (XY):} Another auxiliary task is the prediction of expert-designed targets. It is based on the idea that a good representation should be able to predict key artifacts of an environment. This requires domain knowledge and is not always possible. 
Here we consider the coordinates of the agent in the environment as the target predictions.

\textbf{Virtual Value Function Learning (VirtualVF):} This auxiliary task is based on the tasks the agent will face in transfer. 
We consider one auxiliary loss that uses a goal location at the center of the maze (VirtualVF-1), and another that uses five goals at the four corners and the center of the maze (VirtualVF-5). These are virtual tasks, because the agent imagines achieving these goals, even though they are not the training goal. We use VirtualVF-1 and VirtualVF-5 to assess the utility of having a larger set of virtual goals.
We learn these auxiliary value functions with DQN.

\textbf{Augmented Temporal Contrast (ATC):} 
The contrastive loss encourages the network to learn similar representations for input-states that are temporally close to each other~\cite{stooke2021decoupling,agarwal2021contrastive}. This auxiliary task led to the first successful pre-training of a deep reinforcement learning agent, meaning it led to representations that could be generally reused for other tasks.
ATC also includes other augmentations, like data augmentation. We test it with these additions, to report performance of the originally proposed approach, even though it goes beyond strictly only adding an auxiliary loss.

\section{Experiment Design}~\label{sec:experiment_design}

Our experiments consist of two stages: a representation learning stage in a training task, and a transfer stage using this learned representation in a transfer task. In this section, we outline the details for these two stages in our experiments, and outline the agents and how we evaluate them.

\subsection{Transfer Setup}

The first stage is to train the representation. All representations are trained with a DQN agent in the training task, with goal location depicted in Figure \ref{fg:environments}. To prevent overfitting, we employ an \emph{early-saving} strategy to save the representation function, $\repfunc$, as soon as the agent is able to finish 100 consecutive episodes in 100 steps or less. 
Each representation corresponds to a choice of activation function and auxiliary loss---including choosing not to use an auxiliary loss. 

In the second stage, we learn with the representation from the first stage, in a new transfer task. Specifically, we 1) load and freeze the learned representation, 2) re-initialize the value function and 3) learn the value function for the transfer task with DQN, with the fixed representation. No auxiliary tasks are used in transfer, and only the 64$\times$64 value function network is learned with DQN.  
Learning with a re-initialized value function rather than fine-tuning prevents negative effects from the old value function during transfer, especially to less similar transfer tasks. Further, re-initializing the value function ensures that the difference between transfer learning and learning from scratch is due to the learned representation. The agent learns in this new task for 100 thousand steps. 

We consider 173 transfer tasks -- all possible goal locations, including the training goal state. To sort performance amongst these tasks, we provide a novel method to measure their similarity to the training task. In this way, we can ask questions about transfer to more or less similar transfer tasks. The key idea is to first obtain successor representations for each state, and then compute similarity in this new space. The successor representation encodes similarity based on transition dynamics, meaning that states are considered nearby due to ability to reach them rather than due to other distances, such as Euclidean distances which does not respect the walls in the Maze. For specific details, see Appendix \ref{sec:task_similarity}.

\subsection{Network Choices and Activation Functions}~\label{activation_functions}

To obtain representations with different properties, we use two different activation functions: Rectified Linear Unit (\textbf{ReLU}) \cite{nair2010rectified} and Fuzzy Tiling Activation (\textbf{FTA}) \cite{pan2021fuzzy}.
ReLU is a standard activation function, defined as $\max(z, 0)$ for input $z$, where $z$ is a linear weighting on the previous layer.  

\newcommand{\ftabins}{k}

FTA is a newly introduced activation, designed to generate sparse outputs. Essentially, it bins the scalar input into $\ftabins$ bins, with some smoothing to ensure non-zero gradients through the activation. The smoothness and bin width is controlled by a parameter $\eta > 0$. The interval is from $[-\tfrac{\eta}{2} \ftabins, \tfrac{\eta}{2} \ftabins]$, with $\ftabins$ equally sized bins of size $\eta$. 
Assume the input $z$ is in bin $i$, namely 
$-\tfrac{\eta}{2} \ftabins + (i-1) \eta \le z \le -\tfrac{\eta}{2} \ftabins + i \eta$
where $i \in \{1, 2, \ldots, \ftabins\}$. The $\ftabins$-dimensional output vector $\mathbf{h}(z)$ given by FTA on $z$ has entries $h_j(z) \in [0, 1]$ defined as
\begin{align*}
h_j(z) = \begin{cases}
1 & \text{if } j = i,\\
1 + \eta (j -\tfrac{\ftabins}{2}) - z & \text{if } j < i \text{ and } z > \eta(j -\tfrac{\ftabins}{2}),\\ 
1 + z \!-\! \eta (j\!-\!1 - \tfrac{\ftabins}{2}) & \text{if } j > i, z < \eta (j\!-\!1 - \tfrac{\ftabins}{2}), \\ 
0 & \text{else}.
\end{cases}
\end{align*}
Larger $\eta$ activates more entries in $\mathbf{h}(z)$, and smaller $\eta$ results in more sparsity. This formulation removes a hyperparameter, by using the suggested default choice of $\eta = \delta$.


For our experiments, the representation function consists of two convolutional layers, one linear transformation, and a choice of activation function. The linear layer projects the output of the convolutional layer to a 32-dimensional space. 
When using ReLU, the representation layer has $d =32$ features. If FTA is used, it has 640 features since FTA projects each scalar to a short, sparse vector with 20 bins. Note that FTA still uses the same number of learned parameters to produce this 640 features, as ReLU uses to produce the 32 features, because binning occurs after the linear weighting. However, the outputted number of features is higher, and so the value function and auxiliary tasks all have more parameters, at least in their first layer. We therefore also evaluate ReLU(L)---L for large---which uses 640 features. ReLU(L) uses significantly more parameters to produce these 640 features than FTA. 

The structure for the value function and auxiliary tasks is given in Figure \ref{fg:overview}. We use two hidden layers with 64 nodes each for the value function, and one hidden layer with 64 nodes for the auxiliary task. We use a simpler network for the auxiliary task to force the representation to learn as much as possible. We use a slightly larger network for the value function, to avoid overly constraining it and so confounding transfer performance. 

\subsection{Agent Specification}

We use standard choices for DQN, including the use of $\epsilon$-greedy exploration, an experience replay buffer, target networks, and the Adam optimizer \cite{kingma2014adam}. In total there are 9 choices for auxiliary tasks: No-aux, ATC, IR, NAS, SF, Reward, XY, VirtualVF-1, and VirtualVF-5. There are 3 activations: FTA, ReLU, and ReLU(L). 
When using FTA with auxiliary tasks, we set the number of bins $\ftabins = 20$ and 
$\eta = 0.2$.
This implicitly specifies the range for binning to $[-2, 2]$. For the No-aux task agents, we test $\eta=0.2, 0.4, 0.6,$ and $0.8$ and report performance for each, not the best one. This gives a total of 30 agent specifications. 


We consider three baseline agents, which we call \textsc{Random}, \textsc{Input}, and \textsc{Scratch}. They allow us to falsify different hypotheses about the role of the learned representation. \textsc{Random} uses a randomly initialized network as the representation, without any learning. The agents start with a random network, so this baselines checks whether learning actually improved the representation. \textsc{Input} omits the representation network and directly inputs the agent's observation to the value function component. It is meant to check if the learned representations play any (useful) role, and if  learning from scratch in the transfer task might just have been faster with smaller networks. 
Finally, \textsc{Scratch} is a DQN agent that starts learning from randomly initialized weights in the transfer task. The purpose of learning the representation is to learn faster than learning from scratch in the transfer task. This is the most important baseline, as it defines whether a learned representation was successful for transfer---facilitated learning faster than \textsc{Scratch}---or not---was comparable to or worse than \textsc{Scratch}.

\subsection{Reporting Performance and Hyperparameters}

To report performance, we have to consider how to measure performance and how to set hyperparameters.  
In both the training and transfer tasks, every 10 thousand steps we record the average return of the last 100 episodes. To summarize performance across the 300k steps, we take the sum of these recorded values, also called the Area Under Curve (AUC). The AUC is used to select hyperparameters. 

The only hyperparameter common across all agents is the stepsize; we therefore only sweep this hyperparameter. We separately pick the stepsize in Stage 1, in the training task, and in Stage 2, when just learning the value function. We use the average performance over 5 runs to select the step-sizes. Namely, in Stage 1 we run each of the 30 agent specifications with different step-sizes, for 5 runs. We select the best step-size according to training AUC, and use the representations produced under those step-sizes. Then in Stage 2, we evaluate each stepsize only for those representations, and pick the best step-size for an agent specification for each transfer task by using averaging performance across the 5 runs. We sweep the stepsize to ensure we are evaluating reasonably well-optimized agents.
Additional hyperparameter details, including the selected values, are available in Appendix \ref{app:empiricaldets}.

When we report performance across agents, we do not average across these 5 runs. Instead, each run produces a different representation and we report performance for each one as an independent data point. When showing aggregate performance, we aggregate from this larger pool of 5 runs for each 30 agents specifications, namely over 150 representations. We do so because each representation has different properties; when correlating agent properties and performance, we may not care which auxiliary task was used, but rather only care about its emergent properties. Averaging across runs compares methods (agent specification), rather than representations. 

Finally, we obtain transfer performance in 173 transfer tasks. This means we get 173 transfer performance samples for each of the 150 representations. In total, when aggregating across transfer tasks or agent specifications, we obtain a significant number of samples to estimate aggregate performance, even though each agent specification only has 5 runs in the training task. For example, in Figure \ref{fg:result1}, each bar in the plot is for one agent specification and uses $173\times5 = 865 $ performance samples to estimate medians, means and standard deviations. 
In total, we generate $173 \times 150 = 25,950$ agent-task combinations.

\section{Good, bad and ugly Representations}~\label{sec:design_choices}

We expect some agent specifications to result in representations that aid transfer, and others to impede transfer. \textsc{Unreal}~\cite{jaderberg2016reinforcement}, the first large-scale deep reinforcement learning system to highlight the utility of auxiliary tasks, showed that although auxiliary tasks like pixel prediction improved performance substantially, other tasks such as feature control had a much smaller impact.
Other work has highlighted that it can be difficult to obtain any transfer in reinforcement learning \cite{farebrother2018generalization,tyo2020transferable}. It seems the design and deployment of auxiliary tasks remains largely an art. 

In this section, we provide some clarity on these discrepancies by showing that 1) there is large variability in performance across auxiliary tasks, and 2) transfer performance can degrade significantly as tasks become more dissimilar. 


\begin{figure}
	{\caption{Transfer performance of 105 different representations (ReLU and FTA) on 173 transfer tasks. The tasks on the x-axis are arranged by similarity to training tasks: on the left (small x-values) being most similar and on the right (large x-values) being most dissimilar. The black line shows the performance when learning in each transfer task from scratch. Lines completely above the black line indicate a representation yielded successful transfer in all tasks. Lines that start above the black line but fall below as we move left to right indicate a representation that transfers to similar tasks but not dissimilar tasks. The \textsc{Input} and \textsc{Random} baselines are not competitive; for completeness, we still report their performance, but with ligther lines. Overall, many representations achieve transfer and generally FTA-based representation are better on this problem. Details on how task similarity was computed and how this plot was generated can be found in the Appendix. The transfer performance of ReLU(L) is shown in Figure \ref{fg:auc_relul}; it exhibits the same pattern as ReLU, transferring well to similar tasks and not as well in less similar tasks.}\label{fg:auc_chosen}}
	{\includegraphics[width=0.5\textwidth]{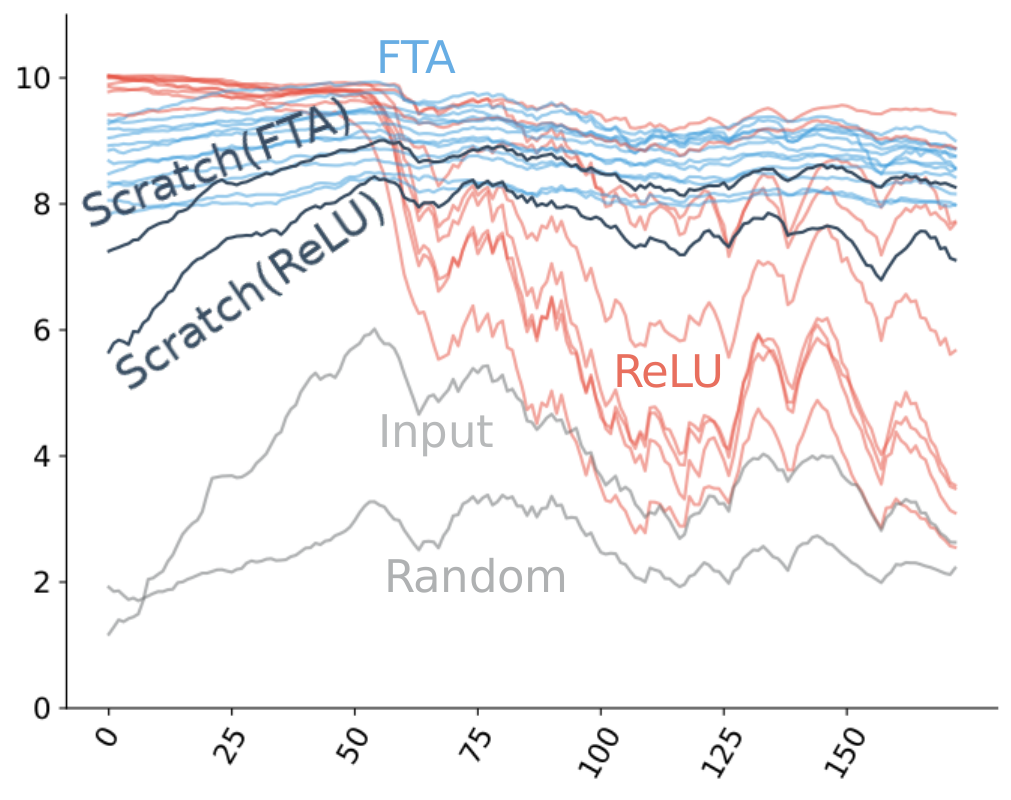}}
\end{figure}

Figure \ref{fg:auc_chosen} summarizes the transfer performance of many different representations corresponding to different auxiliary tasks and activation functions. The plot has task similarity on the x-axis, and each point on the plot summarizes the performance of one representation on one particular transfer task. The lines show how much transfer performance degrades as tasks become more dissimilar.\footnote{Figure \ref{fg:environments} shows two transfer tasks as purple squares. The one beside the training goal is most similar according to our ranking, and the other is least similar to the training goal.} 
The bold black line shows performance in the transfer task if the representation and value function were trained from scratch---no transfer. Any point above the bold black line indicates a representation that achieved better performance than training from scratch on that task---successful transfer. Any line completely above its corresponding black line indicates a representation that achieved successful transfer for all goal states. 

The most important conclusions from Figure \ref{fg:auc_chosen} are that 1)~several representations achieve successful transfer across all tasks, and 2)~a great variety of representations emerge with transfer performance ranging from good to significantly worse than scratch. Looking more closely, some representations achieve successful transfer in dozens of tasks which are most similar to the training tasks, but for tasks less similar to the training one, performance is poor as seen by the step down in many of the lines. 

Generally, we found that FTA-based representations yield better representations for transfer compared to ReLU. Almost all FTA representations outperformed Scratch (FTA), and transfer to less similar tasks was effectively the same as on similar tasks to training, as evidenced by the nearly flat lines in Figure \ref{fg:auc_chosen}. Interestingly, many ReLU-based representations achieved very good performance in transfer to similar tasks but performed significantly worse than Scratch (ReLU)---nearly as bad as the input baseline---on less similar tasks. ReLU (L) performed similarly to ReLU; this result is in Appendix \ref{apdx:relu_l}. Another note of interest is that Scratch (FTA) performs better than Scratch (ReLU), and yet FTA representations were still better able to transfer than ReLU-based ones: using FTA improved on using ReLU and then training the FTA representation first in a training task improved performance in the transfer task even more.

\begin{figure}[!t]
	\centering
	\begin{tikzpicture} 
	\node[rotate=0, align=center, font=\color{black}] {\qquad \quad Total reward, averaged over transfer tasks};
	\end{tikzpicture}
	
	\begin{subfigure}[b]{0.3\linewidth}
		\centering
		\begin{tikzpicture}
		\node[align=center,font=\color{black}] {\quad \footnotesize ReLU};
		\end{tikzpicture}
	\end{subfigure}
	\begin{subfigure}[b]{0.3\linewidth}
		\centering
		\begin{tikzpicture}
		\node[align=center,font=\color{black}] {\quad\;\footnotesize  ReLU(L)};
		\end{tikzpicture}
	\end{subfigure}
	\begin{subfigure}[b]{0.3\linewidth}
		\centering
		\begin{tikzpicture}
		\node[align=center,font=\color{black}] {\quad\;\footnotesize  FTA};
		\end{tikzpicture}
	\end{subfigure}
	
	\begin{subfigure}[b]{0.3\linewidth}
		\begin{tikzpicture}
		\node (img)  {\includegraphics[width=\linewidth]{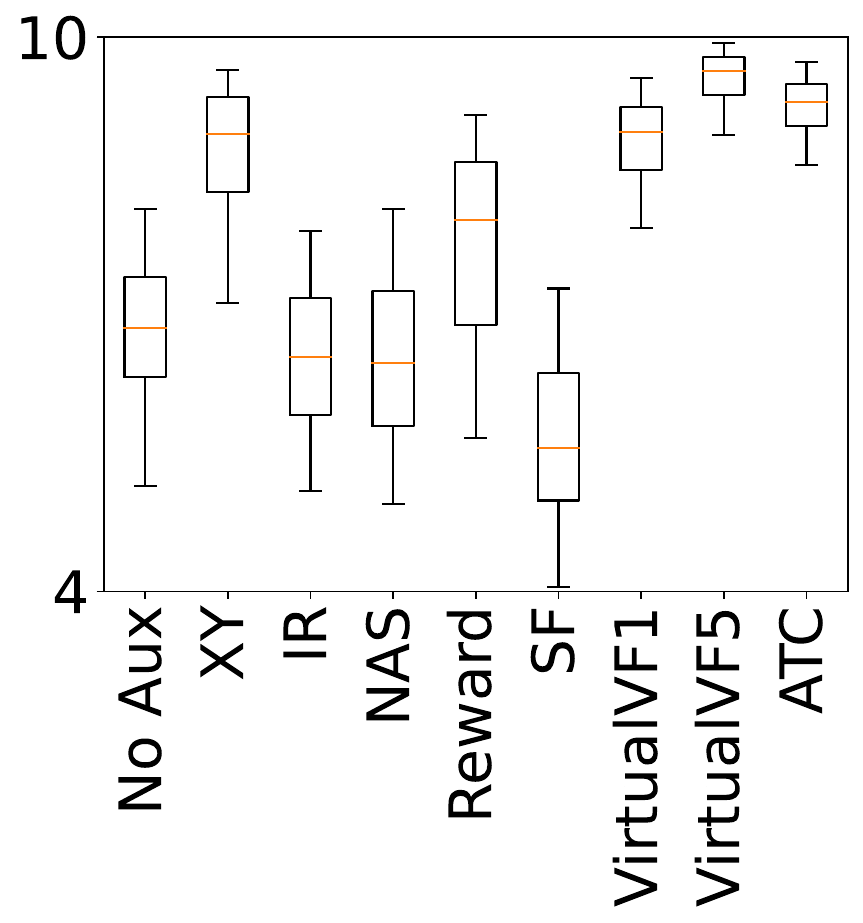}};
		\end{tikzpicture}
	\end{subfigure} 
	\begin{subfigure}[b]{0.3\linewidth}
		\begin{tikzpicture}
		\node (img)  {\includegraphics[width=\linewidth]{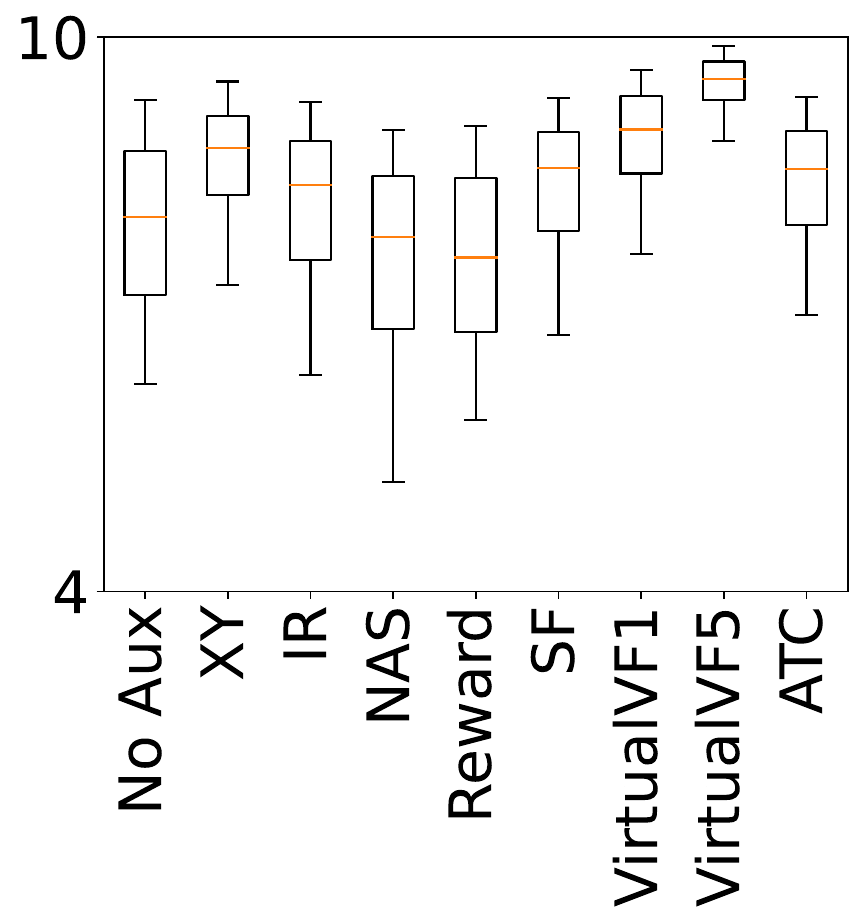}};
		\end{tikzpicture}
	\end{subfigure}
	\begin{subfigure}[b]{0.3\linewidth}
		\begin{tikzpicture}
		\node (img)  {\includegraphics[width=\linewidth]{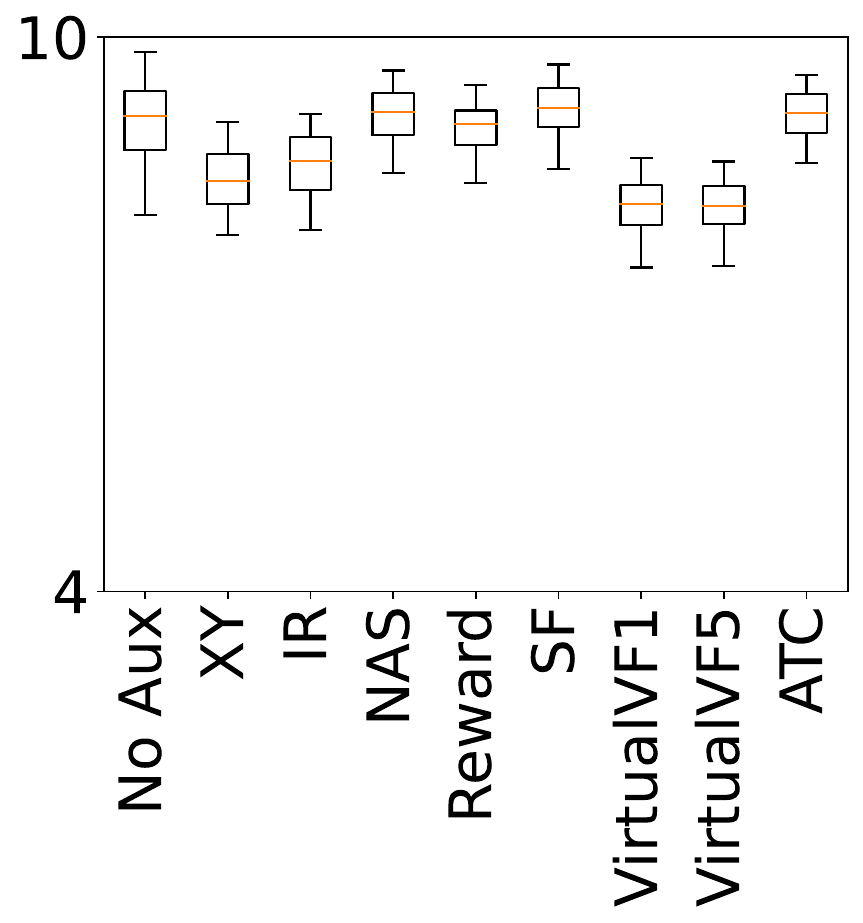}};
		\end{tikzpicture}
	\end{subfigure}
	
	\caption{Transfer performance depends on the activation function, representation size, and auxiliary tasks. Overall, FTA-based representations achieved the best performance and exhibited the least variation in performance across auxiliary tasks. The orange lines depict the median, the upper and lower edges of the box show the $25_{th}$ and the $75_{th}$ percentiles, while the whiskers show 1.5 times the inter-quartile range. These results are computed over $173 \times 5 = 865$ samples, and so the standard errors are quite small (as you can see in Figure \ref{fg:atc_ac_activation_95CI} in Appendix \ref{app_ci}). }
	\label{fg:atc_ac_activation}
\end{figure}
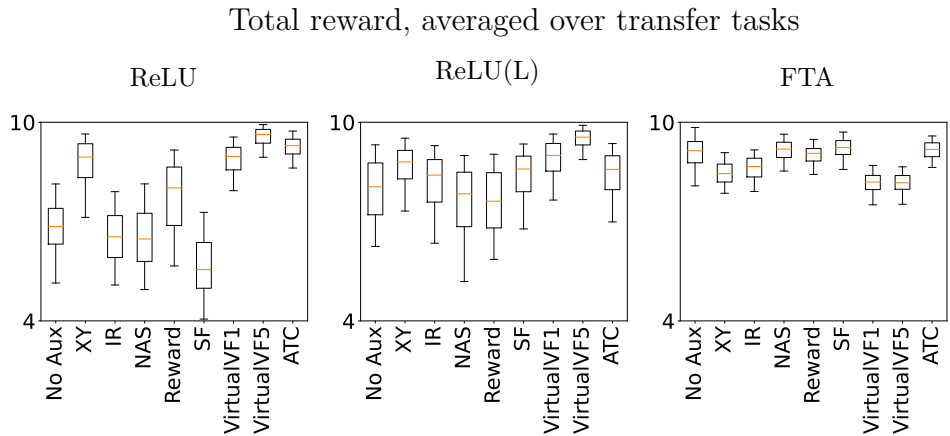

Digging a little deeper, Figure~\ref{fg:atc_ac_activation} depicts the transfer performance of each auxiliary task. We again see that FTA-based representations achieve higher performance overall and higher performance across auxiliary tasks---the worst performing representation never used FTA, always ReLU. Inspecting each auxiliary task, the FTA-based representations exhibited lower variance across runs.  Larger ReLU representations, ReLU(L), did improve performance over the smaller ReLU representations, but not uniformly. The IR auxiliary task representation, for example, improves with large ReLU networks, but ATC performs worse---though not significantly in either case.

At the auxiliary task level, there are no obvious trends (except for the fact IR and Reward are generally not useful). For example, the successor-feature auxiliary task (labelled SF) is among the best performing FTA representations and among the worst performing ReLU representations. 
The subgoal-navigation auxiliary tasks (VirtualVF) result in the best performance with ReLU representations. These subgoals can be thought of as way-points placed at strategic locations in the environment; perhaps these tasks force the network to represent how to navigate to these waypoints which then speeds learning when navigating to other nearby goals in transfer. Perplexingly, these are not the best performing representations when combined with FTA activation functions. Perhaps FTA networks already extract a general and transferable representation (as evidenced by the performance of `No Aux'), and thus the subgoal auxiliary tasks simply do not help much. It is difficult to know looking at performance only; in the following sections we look at different properties of the learned representations as a lens to understand such mysteries. \looseness=-1   


We included ATC, a recent representation learning strategy, to calibrate the quality of transfer performance. This approach uses multiple networks to compute a contrastive loss, while also using data augmentation of the input images. ATC worked well, but it did not significantly outperform the best ReLU and FTA representations. In addition, we found that the ReLU network combined with data augmentation (random shifting of the input imagines) and no contrastive setup achieved similar performance to ReLU ATC.

\section{Representational Properties, Old \& New} \label{sec:properties}


Before the emergence of deep reinforcement learning, the study of representations in reinforcement learning and their effects on learning was focused on fixed bases. The problem with this is not that a fixed basis cannot capture complex non-linear relationships (e.g., see the work by Liang et al.~\cite{liang2016state}), but rather that the representation is fixed---the features are not adapted to the task. In some sense, this is good because it forces the agent designer to consider what are desirable representation properties---a level of analysis complementary to the design of good algorithms.
Over the years, researchers have proposed and debated numerous properties. We leverage these discussions to analyze our learned representations.

We characterize a representation into three main axes: \emph{capacity}, \emph{efficiency}, and \emph{robustness}. Capacity reflects whether a representation can represent a given function. Efficiency captures the lack of redundancy of the features and the computational cost of using them. Robustness captures the idea that interference is undesirable and that representations should avoid it; a more complete name is \emph{update robustness}. We define six metrics that capture these three axes, and we use them to evaluate the representations learned by our agents. 
Our goal is to develop a systematic methodology for assessing learned representations, based on a diverse set of properties. This evaluation list does not suggest that a property is necessary; rather, it provides some quantitative measures to supplement more qualitative evaluation like visualization. Such a list is necessarily incomplete; we attempt only to start with a reasonably broad set of properties. 

We assume we have a dataset of 1000 transitions to measure the properties, $\mathcal{D}_\text{test} = \{(s_1, a_1, s_1', r_1), (s_2, a_2, s_2', r_2),$ $\ldots, (s_N, a_N, s_N', r_N)\}$, where $s_i'$ is the outcome state from $s_i$. This dataset is obtained by running the random policy for $N$ episodes, and then randomly subsampling $N$ transitions, to ensure we cover the state space. We store the transition because some of the properties rely on consecutive states or the entire transition. The symbol $\phi_i$ refers to the representation of $s_i$; $V(\phi) = \max_a Q(\phi_i, a)$ is the value learned given that representation. We compute distances both according to the representation and according to action-values,
\begin{equation}
\begin{aligned}
d_{v,i,j} &\defeq |V(\phi_i) - V(\phi_j)|, \\
d_{s,i,j} &\defeq \|\phi_i - \phi_j\|_2.
\end{aligned}
\label{eq_dist}
\end{equation} 

\subsection{Capacity: Retaining Relevant Information and\\ Nonlinear Transformations}

The first property to consider for a representation is its \textbf{capacity}: can it represent the functions we want to learn? The value function network should be a simple function of these features, such as a simple neural network.
To measure capacity, we use one direct measure, \emph{complexity reduction}, and two indirect measures, \emph{dynamics-awareness} and \emph{diversity}.

\textbf{Complexity reduction} reflects how much the representation facilitates simplicity of the learned value function on top of those features. If complexity is small, the features encode much of the non-linearity needed. A well-known result is that the composition of a Lipschitz function $V$ with another function $\phi$ has (Rademacher) complexity: complexity($V \circ \phi$) $\le L$ complexity($\phi$), for $L$ the Lipschitz constant of $V$. The Lipschitz constant $L$ is one where $\tfrac{d_{v,i,j}}{d_{s,i,j}} \le L$ for any $\phi_i, \phi_j$. 
For a smaller $L$, the representation $\phi$ handles most of the complexity, which is good because we have a longer initial learning phase to obtain $\phi$. $V$ on top of $\phi$ should be fast to learn.  
Lipschitz value functions have also been motivated for value transfer \cite{lecarpentier2021lipschitz} and learning models \cite{farahmand2017value}. 

Let us turn this intuition into a measure of how much the representation helps reduce the required complexity in the value function. We can measure $\tfrac{d_{v,i,j}}{d_{s,i,j}}$ for all pairs $(i,j)$ in our dataset.  An estimate of $L$ is the maximum over these slopes. However, empirically we find that the average of these slopes results in bigger differences between representations with better correlation to transfer performance. The max loses significant precision, possibly making very different $Q$ have similar measures; the average summarizes the whole surface. For example, one representation might reduce complexity for most of the space, but due to one small part, have a similarly high maximum slope to a representation that barely reduces complexity in any part of the space.  
We call these averaged ratios $L_{\text{rep}}$, giving 
%
\begin{align}
\text{Complexity Reduction} &\defeq 1 - \frac{L_{\text{rep}}}{L_\text{max}} \label{fml:complexy}\\
\text{where } \quad L_{\text{rep}} &\defeq \tfrac{2}{N(N-1)}{\textstyle \sum^N_{i,j,i< j}} \tfrac{d_{v,i,j}}{d_{s,i,j}} \nonumber
.
\end{align}
When this ratio is computed on given time step $t$---either during learning or on the last time step before the representation is frozen for transfer---we use the current action-values.
We normalize $L_{\text{rep}}$ between 0 and 1 using $L_\text{max}$, computed as the maximum $L_{\text{rep}}$ over all representations and across time steps. This is subtracted from 1 to ensure higher values refer to higher reduction in complexity.  

We can also indirectly measure complexity---that is without specifying a set of value functions---by testing if the representation is \textbf{dynamics-aware}. This means that pairs of states where one is a successor to the other should have similar representations, and states further apart in terms of reachability should have a low similarity. This measure is in fact related to the Laplacian used for proto-value functions and successor features \cite{stachenfeld2014design,machado2018eigenoption}.
For every state in the dataset, we take its successor state and a random state. If the distance, in representation space, between the successor state is smaller than the distance to a random state, then the representation has high dynamics awareness.
\begin{equation}
\!\!\text{Dynamics Awareness}\!\defeq\!\tfrac{\sum^N_{i}||\phi_i - \phi_{j\sim U(1,N)}|| - \sum^N_{i} ||\phi_i-\phi'_i||}{\sum^N_{i}||\phi_i - \phi_{j\sim U(1,N)}||}.
\label{fml:dynamics}
\end{equation}

In addition, we can measure the \textbf{diversity} of a representation, which is the opposite of \emph{specialization}. If a representation is specialized to one value function, then it likely uses a small subspace of the larger Euclidean space and likely does not produce a diversity of possible feature vectors. This specialization may be problematic, as it means the representation is unlikely to perform well when it is transferred to learn another value function. 

To define diversity, we use a ratio between state and value differences. Given two states $s_i$ and $s_j$, we can compare the distance between their representations ($d_{s, i, j}$) and the distance between their values ($d_{v, i, j}$).
If the value distance is high---the two state values are very different---then the representation distance is also likely to be high to allow this. The interesting case is when the value distance is low. The representation distance can be high or low, and still allow two states to have similar values, because we project from a higher-dimensional feature vector to a scalar value. A representation with high diversity would have high representation distance when possible, allowing two states to be distinguished even when they have similar values. 
A representation with low diversity would simply map these two states with similar values to similar representations, specializing to this value function. 
The measure is
\begin{equation}
\text{Diversity} \!\defeq\! 1 \!-\! \tfrac{1}{N^2} \!\!\sum^N_{i,j}\min{\!\left(\frac{d_{v,i,j} / {\scriptstyle \max_{i,j} d_{v,i,j}}}{d_{s,i,j} / {\scriptstyle \max_{i,j} d_{s,i,j}} \!+\! 10^{-2}}, 1\right)} \label{fml:diversity}
.
\end{equation}
We normalize by the maximum distances, to be invariant to value and representation scales. Diversity can be seen as 1-specialization. The specialization is lower when $d_{v,i,j}$ is small and $d_{s,i,j}$ is large, causing this ratio to be closer to zero. The specialization is higher when the ratio between $d_{v,i,j}$ and $d_{s,i,j}$ is nearly one.
Diversity allows us to indirectly measure capacity, as we can check the level of specialization for a given function without needing to have access to the larger set of possible functions.

\subsection{Efficiency: Feature Redundancy}

Many function classes can satisfy these capacity properties and so we consider other functional properties of the features. Reducing redundancy in the representation, finding \emph{linearly independent} features, is a basic requirement. \textbf{Orthogonality} satisfies this requirement and additionally provides distributed features as well as minimal interference. For example, factor analysis finds a dense set of orthogonal (latent) factors to explain the data. This representation is highly distributed, as each feature is used to describe many different inputs. 
At the same time, interference is reduced: the interference for two states with orthogonal feature vectors is zero under linear updating. 
As before, we normalize magnitudes and ensure higher orthogonality means that more feature vectors $\phi_i$ and $\phi_j$ are orthogonal to each other. 
\begin{equation}
\text{Orthogonality} \defeq 1 - \tfrac{2}{N(N-1)}\sum^N_{i,j,i< j} \frac{|\langle \phi_i, \phi_j \rangle|}{\|\phi_i\|_2 \|\phi_j \|_2}.
\label{fml:orthogonality}
\end{equation}
Note that there is an equivalence between orthogonal feature vectors ---orthogonal representations---and orthogonal features: the sum over all states $i,j$ of $\langle \phi_i, \phi_j \rangle^2$ is equal to the sum over all pairs of features of the dot product between the vector of those feature values across states (see Appendix \ref{app_orth_relationship}). Additionally, for centered features, orthogonality is also equivalent to decorrelation (see Appendix \ref{app_orth_correlation}). 

One idea related to orthogonality is \textbf{sparsity}. If only a small number of features are active for an input, then the features are {sparse}---with typically the additional condition that each feature is active for some inputs (no dead features). For non-negative features, maximizing sparsity corresponds to finding orthogonal features: dot products can only be zero when features are non-overlapping for two inputs.
Sparsity has the additional benefit, though, of improving efficiency for querying and updating the function, because only a small number of features are active. 
To measure sparsity, we calculate the percentage of inactive features on average across states in the dataset. 
\begin{equation}
\text{Sparsity} \defeq \frac{1}{d N} \sum_{i=1}^N \sum^d_{j=1} \mathds{1}(\phi_{i,j}=0),
\label{fml:sparsity} 
\end{equation}
where the representation $\phi_i$ for state $s_i$ is $d$ dimensional. Equality with zero is tested within a tolerance of $10^{-10}$. 

\subsection{Robustness: Interference Reduction}

More recent work in neural networks has also focused on {robustness}, both to interference and noise. \textbf{Interference} reflects how much updates in one state reduce accuracy in other states. We use a recent measure developed for reinforcement learning \cite{liu2020towards}, which uses the difference in temporal difference errors before and after an update. We do the comparison each time the target network is synchronized, which occurs every 64 steps, for a total of $T$ times during learning. For every $t = 1, \ldots, T$, we compare the error between $\theta_t$ and the parameters after 64 updates, $\theta_{t+1}$; note $t$ here references the synchronization iterator rather than time. 
\begin{align}
&\text{Non-interference} \defeq 1 - \frac{\text{Interference}}{\text{MaxInterference}} \label{fml:noninterf}, \\
&\text{Interference} \defeq \{\text{Update Interference}_{t, \geq Percentile_{0.9}}\}_{t=1}^T, \nonumber\\
&\text{Update Interference}_{t} \defeq \frac{1}{N} \sum_{i=1}^N \text{err}_{t,i}(\theta_{t+1}) - \text{err}_{t,i}(\theta_{t}), \nonumber\\
&\text{err}_{t,i}(\theta) \defeq \left(r_{i+1}+\gamma_t \max_a Q_{{\theta}_t}(s_{i}', a) - Q_{\theta}(s_i, a_i)\right)^2. \nonumber
\end{align}
The maximal Interference is across all representations.

\begin{figure*}[t]
	\centering
	\begin{subfigure}[b]{0.09\linewidth}
		\centering
		\begin{tikzpicture}
		\node[rotate=0, align=center,font=\color{black}] {\scriptsize Total \\\scriptsize reward, \\\scriptsize averaged \\\scriptsize over \\\scriptsize transfer \\\scriptsize tasks \vspace{0.8cm}};
		\end{tikzpicture}
	\end{subfigure}
	\begin{subfigure}[b]{0.78\linewidth}
		\begin{tikzpicture}
		\node (img)  {\includegraphics[width=\linewidth]{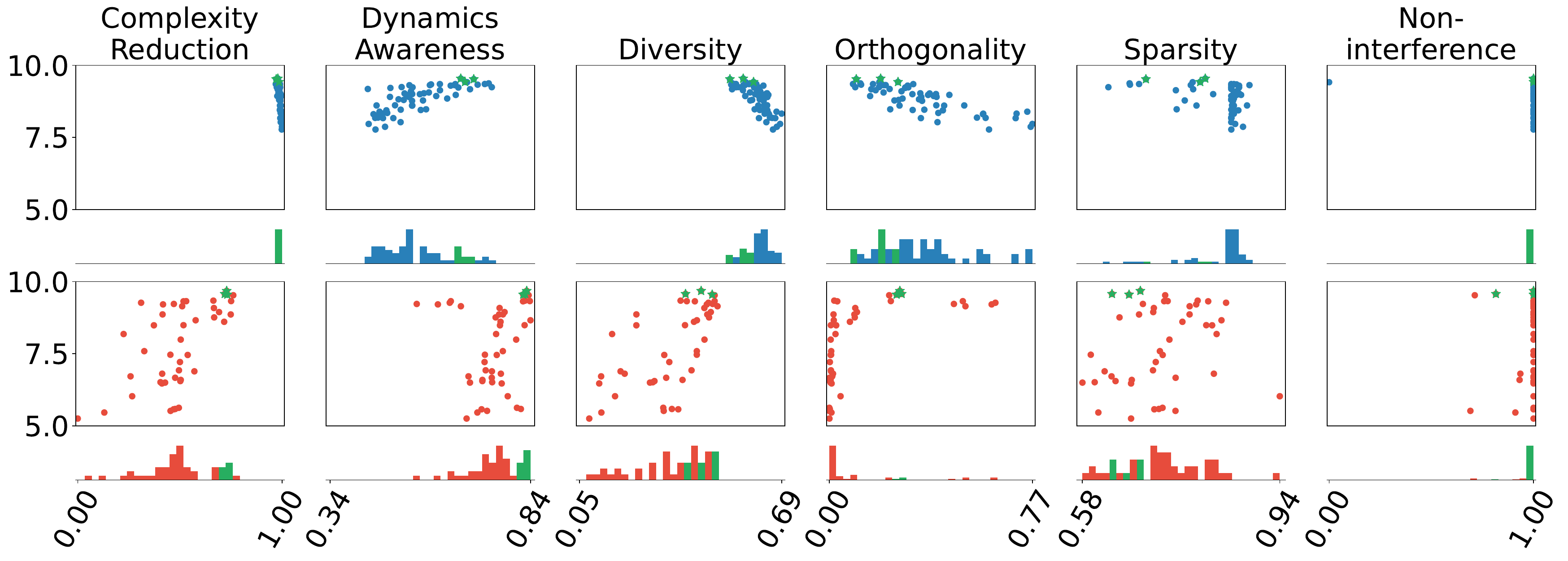}};
		\end{tikzpicture}
	\end{subfigure}
	\begin{subfigure}[b]{0.09\linewidth}
		\begin{tikzpicture}
		\node[rotate=0, align=center,font=\color{black}] {\scriptsize FTA \\\vspace{0.6cm}\\\scriptsize ReLU\vspace{1.2cm}};
		\end{tikzpicture}
	\end{subfigure}
	\caption{Performance averaged over transfer tasks versus representation property values. Each dot in a plot corresponds to one representation, at the (x,y) point corresponding to its property value and average transfer performance. The histograms under the plots visualize the density of points at each property value. The green stars correspond to the three best performing representations. We separate out FTA-base and ReLU-based representation, which exhibit notably different behavior. 
	}
	\label{fg:accum_all}
\end{figure*}
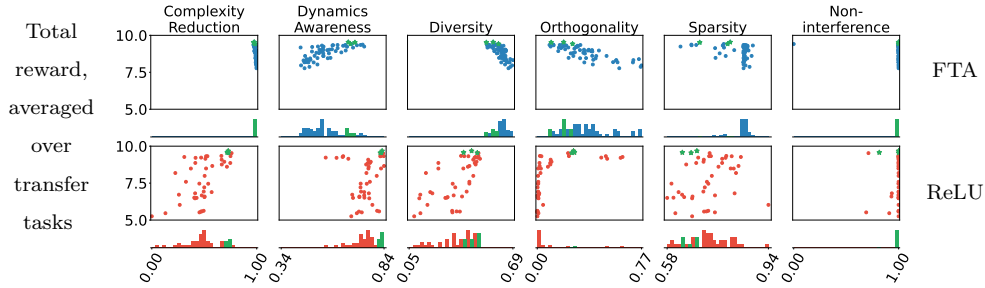
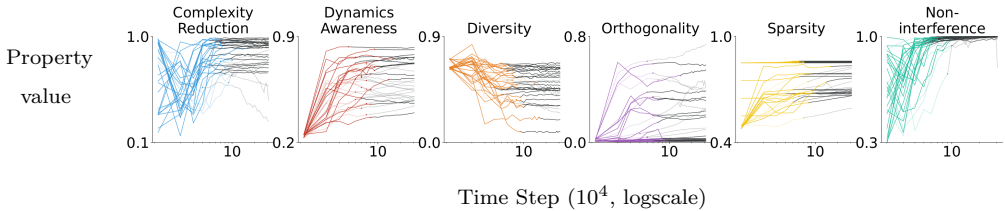
\begin{figure*}[t]
	\vspace{-0.4cm}
	\centering
	\begin{subfigure}[b]{0.03\linewidth}
		\begin{tikzpicture}
		\hspace{-0.5cm}
		\node[rotate=0, align=center,font=\color{black}] {\scriptsize Property \\ \scriptsize value \vspace{0.8cm}};         
		\end{tikzpicture}
	\end{subfigure}
	\begin{subfigure}[b]{0.95\linewidth}
		\centering
		\begin{tikzpicture}
		\node (img)  {\includegraphics[width=0.9\linewidth]{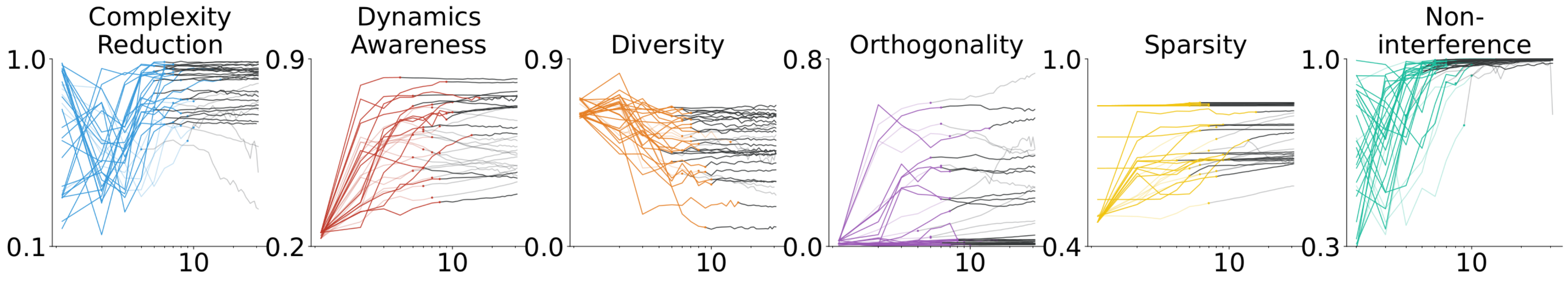}};
		\end{tikzpicture}
	\end{subfigure}
	
	\begin{subfigure}[b]{0.1\linewidth}
		\begin{tikzpicture}
		\hspace{-0.6cm}
		\node[rotate=0, align=center,font=\color{black}] {\scriptsize Time Step ($10^4$, logscale)};
		\end{tikzpicture}
	\end{subfigure}
	
	\caption{
		Plotting the representation properties over time, with one subplot per property. 
		Each curve shows the property of one agent specification (activation and auxiliary task pair), averaged over the 5 runs in the training task. The curve changes color, to black, at the time point where we took the representation and fixed it; this point was chosen based on when the return for the agent stopped changing. Line colors vary from light gray to black based on how much their value fluctuate from the property value after their return converges, with darker lines denoting lower variation. In these plots, we allowed the representation to keep learning to understand if properties significantly change afterwards. Our primary focus is to show the general trend that properties converge over time, and that they converge approximately when the return does; therefore we use the same color for all agent specifications. 
	}
	\label{fg:property_change}
\end{figure*}

\section{The Properties of Good Representations}~\label{sec:properties_results}

We can now return to the main question of this work: do good representations that facilitate transfer exhibit particular properties? In this section, we investigate how the properties defined in the previous section relate to transfer performance. 

Figure \ref{fg:accum_all} contains the main result; for now let us focus on only the FTA representations in the top row to better understand this figure. Each subplot shows the transfer performance of every single FTA-representation averaged over all transfer tasks. The representations in each subplot are ordered based on a single measured property. For example, the first subplot in the first row plots the transfer performance of FTA representations and the dots are ordered by {\em complexity reduction} on the x-axis. Representations with high complexity reduction and good transfer performance would appear as a dot in the top right of the subplot. Representations with low complexity reduction and good transfer performance would appear as a dot in the top left of the subplot, and so on.  

At the highest level we see FTA (top row) and ReLU (bottom row) exhibit different properties across representations. FTA-based representations by large exhibit high complexity reduction and high diversity, whereas ReLU representations range widely from low to medium on the same two measures. In fact, the {\em lowest} observed complexity reduction and high diversity of any FTA representation was greater than the {\em highest} observed complexity reduction and high diversity for ReLU. ReLU representations could be sparse and have low or high orthogonality, whereas FTA representations are mostly sparse. Interestingly, the top representations in terms of sparsity were ReLU. ReLU representations with similar property values can achieve very different transfer performance (visible as points stacked vertically). There appears to be no clear relationship between sparsity and performance for ReLU representations.

Now consider the properties of the top performing representations. Again, let us focus our attention on the FTA representations in the top row of Figure \ref{fg:accum_all}. The green stars in each subplot correspond to the top performing representations (in terms of transfer). First notice the stars are typically close together in x and y indicating all three achieve similar performance with similar property values; this is true for ReLU representations as well. In general, we see that the best performing representations are not at the extremes of any property (high or low). Given that FTA representations by large exhibit high complexity reduction, diversity and sparsity, it is notable that the best performing representations are the lowest of those three properties. 

\emph{In general, the best representations for both FTA and ReLU exhibit fairly similar properties} (relative to other representations with the same activation function): high complexity reduction, medium-high dynamic awareness, and medium orthogonality and sparsity.
Of particular note is the clear pattern in complexity reduction and diversity for ReLU: both needed to be higher, and performance clearly drops for lower values. FTA seems to more naturally produce representations that are higher on these measures; we hypothesize that this is the main explanation for why FTA representations work well across the board for transfer.

The property values depicted in Figure~\ref{fg:accum_all} were computed from the representations when frozen for transfer, but one might wonder what are the dynamics of the properties over time. Recall that we froze and transferred each representation after 100 episodes were completed in the training phase. This choice balances the need for reasonable performance without having to select somewhat arbitrary steps budgets or performance criteria. However, our choice does mean that each representation could receive different amounts of experience, and thus begs the natural question: would the properties reported in Figure \ref{fg:accum_all} be very different with more or less training. Figure \ref{fg:property_change} provides the answers.

Generally, across all auxiliary tasks and activation functions, the representation property values remained similar after initial transients in early learning. Each subplot of Figure \ref{fg:property_change} shows a particular property value for every single representation tested over an extended training period. We intentionally do not distinguish between activation functions and auxiliary tasks in this plot. The change in color indicates when the representation was frozen for transfer, in terms of training time. Note, many representations were frozen after the same number of training steps. Orthogonality, dynamic awareness, and sparsity of a small number of representations slowly increases with more training and complexity reduction of a few representations slowly decreases. Overall, the properties for the most part converge, and do so just before the representations were frozen for transfer. Training the representations longer would not have resulted in significant changes to the property values.

\begin{figure}[t!]
	\centering
	\begin{subfigure}[b]{0.35\linewidth}
		\begin{tikzpicture}
		\node (img)  {\includegraphics[width=\linewidth]{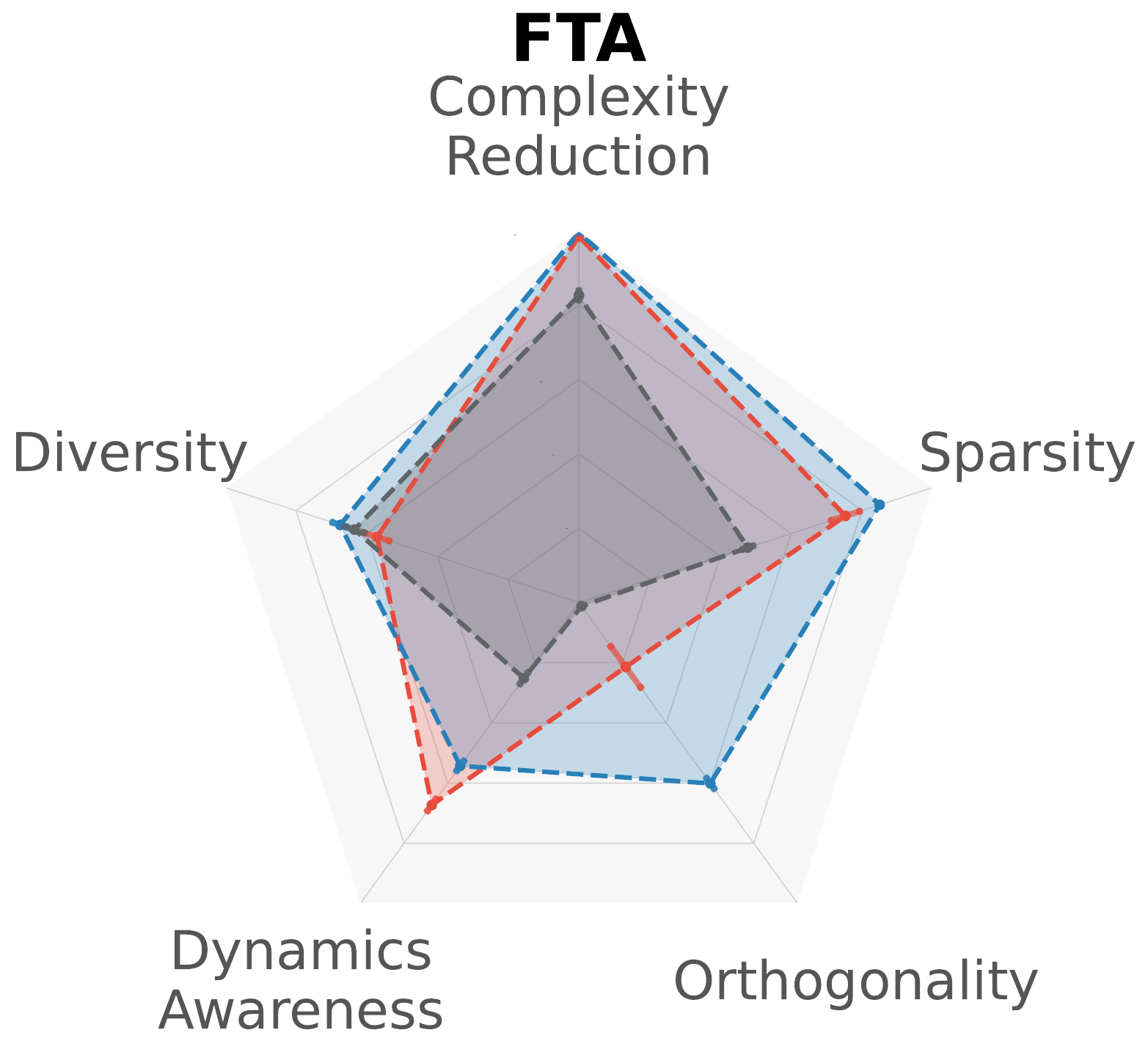}};
		\end{tikzpicture}
	\end{subfigure}   
	\begin{subfigure}[b]{0.35\linewidth}
		\begin{tikzpicture}
		\node (img)  {\includegraphics[width=\linewidth]{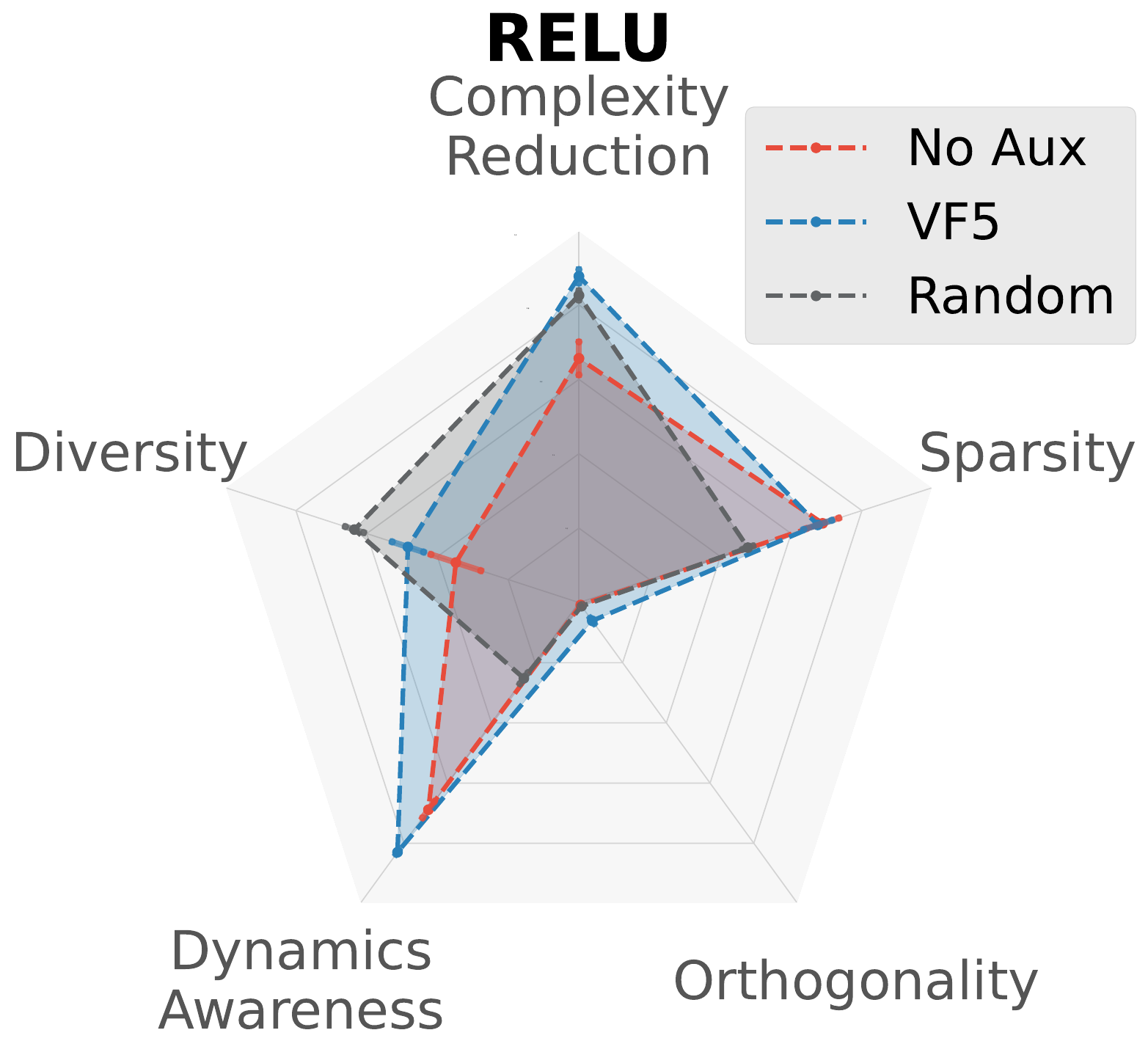}};
		\end{tikzpicture}
	\end{subfigure} 
	
	\vspace{-.25cm}
	\caption{The VirtualVF5 task produces bad FTA representations but improves ReLU representations. Each subplot shows a property value achieved by four different representations: FTA and ReLU with VirtualVF5 and FTA and ReLU with no auxiliary tasks. It is clear this auxiliary task changes the properties of the representations; particularly Dynamics Awareness and Orthogonality. We did not include non-interference as VirtualVF5 had no impact on it.
		\label{fg:vf5}}
\end{figure}

Finally, we investigated why VirtualVF5 was helpful for ReLU and harmful for FTA through the lens of our representation properties. 
We plotted properties for these representations in Figure \ref{fg:vf5}, and found that the addition of this auxiliary loss significantly decreased dynamics awareness for FTA---to the detriment of performance---but increased it for ReLU. Additionally, it caused the FTA-based representation to have much higher orthogonality, likely increasing it to one of the extremes that performed more poorly. For ReLU, the increase in orthogonality was to an interim level, from a value that was very small. It is yet unclear why this auxiliary loss caused these effects, but this clear and systematic change in the properties helps explain this outcome. 

\section{Transfer across Atari 2600 Game Modes}\label{sec_atari}

\begin{figure}[t]
	\centering
	\centering
	\includegraphics[width=0.8\columnwidth]{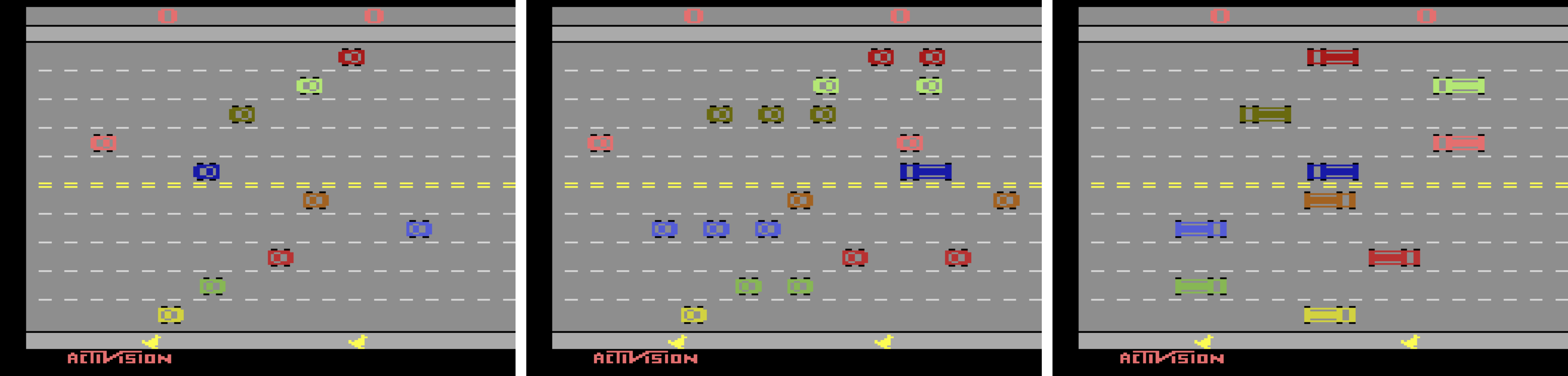}
	\caption{Three different modes of the Atari game \textsc{Freeway}. 
		Atari 2600 cartridges allowed players to choose between game variations that modified sprites, velocities, etc. These different modes are an excellent platform for investigating transfer. They are free from experimenter's bias and, because of hardware limitations, the different modes of a game were often small variations of the default mode, thus are not too different from one another. In this particular example, in \textsc{Freeway}, the default mode (left is m0) consists of a thin traffic in which there are only cars and these cars move at a relatively slow speed. The mode in the center (m1) has trucks, traffic is heavier and moves much faster. The mode on the right (m3) has trucks in all lanes with varying speeds.}
	\label{fg:atari_modes}
\end{figure}

In this section we investigate our methodology and properties on a larger environment, with a substantially more complicated and high-performance agent.
So far, in order to perform a thorough empirical evaluation, our analysis focused on a small, controllable pixel-based navigation environment. 
To demonstrate the generality of our approach, we apply our methodology to understand representations learned for transfer across different Atari 2600 game modes~\cite{bellemare2013the,machado2018revisiting} (see Figure~\ref{fg:atari_modes}). We chose this setting because transfer in Atari is an active research question, with some recent promising results~\cite{rusu2022probing} when transferring policies learned by a state-of-the-art agent called Rainbow~\cite{hessel2018rainbow}. 
We used the same three Atari games: Freeway, Space Invaders and Breakout.


Rainbow can be seen as an auxiliary task learning agent, with auxiliary heads produced by two value heads (duelling networks~\cite{wang2016dueling}) and estimates of the distribution of returns (C51~\cite{bellemare2017distributional}). 
More agent details are given in Appendix \ref{app_atari}. 


The experimental protocol is very similar to before.
We still learn in one source task (the default game mode 0), where the agent learns for 200 million frames, with the learned policy evaluated every iteration for 500,000 steps. In the target task (a different game mode), the representation is frozen and the agent learns the values from scratch for 10 million frames. 
We contrast the representations and transfer performance of Rainbow to a simple baseline: a random representation. We take the exact same Rainbow agent, with the same network architecture, and fix the representation to its random initialization. The value function is still updated in the same way, using dueling networks, distributional updates, prioritized replay and so on. 

Previous transfer results \cite{rusu2022probing} investigated transferring both the learned values and the representation and allowed fine-tuning. Rusu et al.~\cite{rusu2022probing} found Rainbow policies can transfer. We are asking a different question: can the learned representation transfer and what are its properties? 

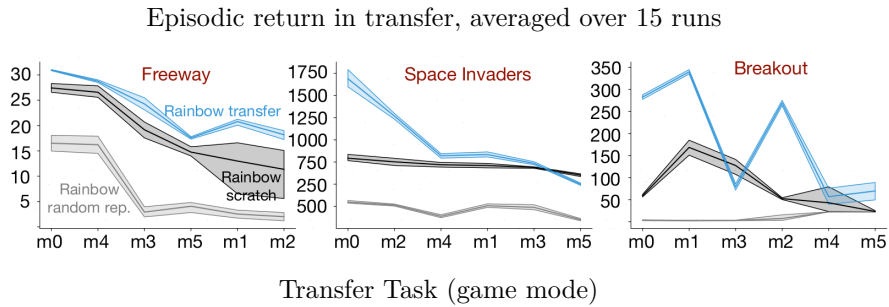
\begin{figure}[t]
	\centering
	\begin{subfigure}[b]{\linewidth}
		\centering
		\begin{tikzpicture}
		\node[rotate=0, align=center,font=\color{black}] {\footnotesize Episodic return in transfer, averaged over 15 runs };
		\end{tikzpicture}
	\end{subfigure}
	\begin{subfigure}[b]{0.85\linewidth}
		\begin{tikzpicture}
		\node (img)  {\includegraphics[width=\linewidth]{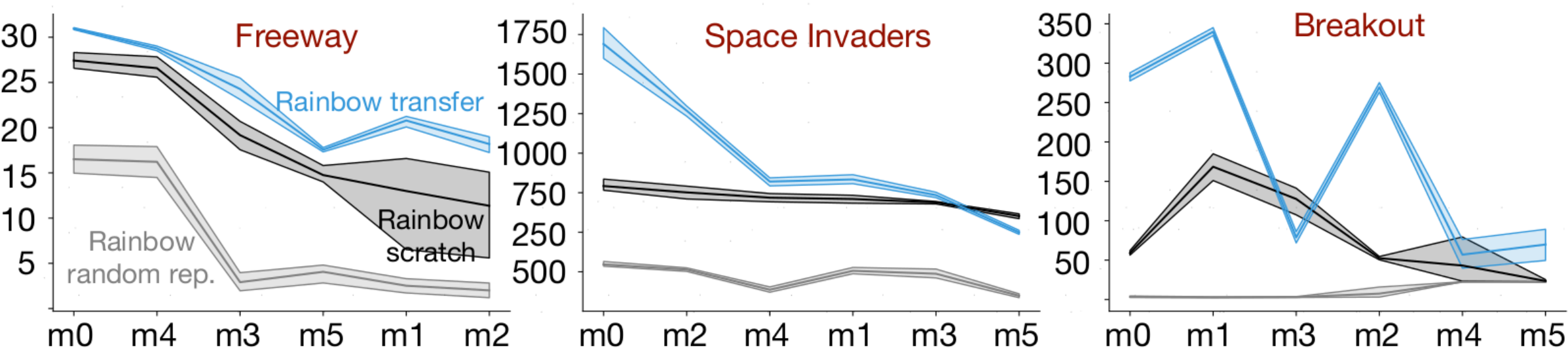}};
		\end{tikzpicture}
	\end{subfigure}
	\begin{subfigure}[b]{\linewidth}
		\centering
		\begin{tikzpicture}
		\node[rotate=0, align=center,font=\color{black}] {\footnotesize Transfer Task (game mode)};
		\end{tikzpicture}
	\end{subfigure}
	
	\caption{
		Transfer in three Atari games, to game modes of increasing difficulty. 
		The episodic return (computed over 500,000 steps) of the greedy policy is obtained every 1 million training steps, then averaged into a single performance number (AUC), plotted on the y-axis. Shaded regions correspond to standard 95\% confidence intervals.  
		The source task, default game mode zero, is plotted first. The remaining modes are ordered by the AUC achieved by training rainbow from scratch. In some cases, the default game mode appears to be more difficult, making performance not monotonically decreasing. Some of the gaps between Rainbow and Scratch look small, but they are actually substantial. They look smaller only because Random baseline does so poorly, skewing the scale of the plot. Notice that in Freeway, the representation even transferred to quite different modes (e.g., m1 and m3 visualized in Figure \ref{fg:atari_modes}), with heavier traffic, different vehicle speeds and vehicle types (trucks). \looseness=-1
	}
	\label{fg:atari_transfer}
\end{figure}

We first look at the transfer performance of Rainbow and Rainbow with a fixed random representation, in comparison to learning from scratch. In Figure \ref{fg:atari_transfer}, we can see two clear outcomes. In general, we see successful transfer: the initial learning phase in the source task resulted in improved performance for Rainbow, as compared to learning from scratch (if the blue line is above the black line positive representation transfer is observed). It was not entirely obvious that this positive representation transfer would be possible, as prior work transferred the entire network and utilized fine-tuning. Second, the random representation generally does not allow the agent to learn in the target task, indicating learning features in the source mode facilitated transfer. This result is surprisingly similar to our results with DQN in the maze (compare Figure \ref{fg:auc_chosen} and Figure \ref{fg:atari_transfer}).  

\begin{figure*}[t!]
	\centering
	\begin{subfigure}[b]{0.32\linewidth}
		\begin{tikzpicture}
		\node (img)  {\includegraphics[width=\linewidth]{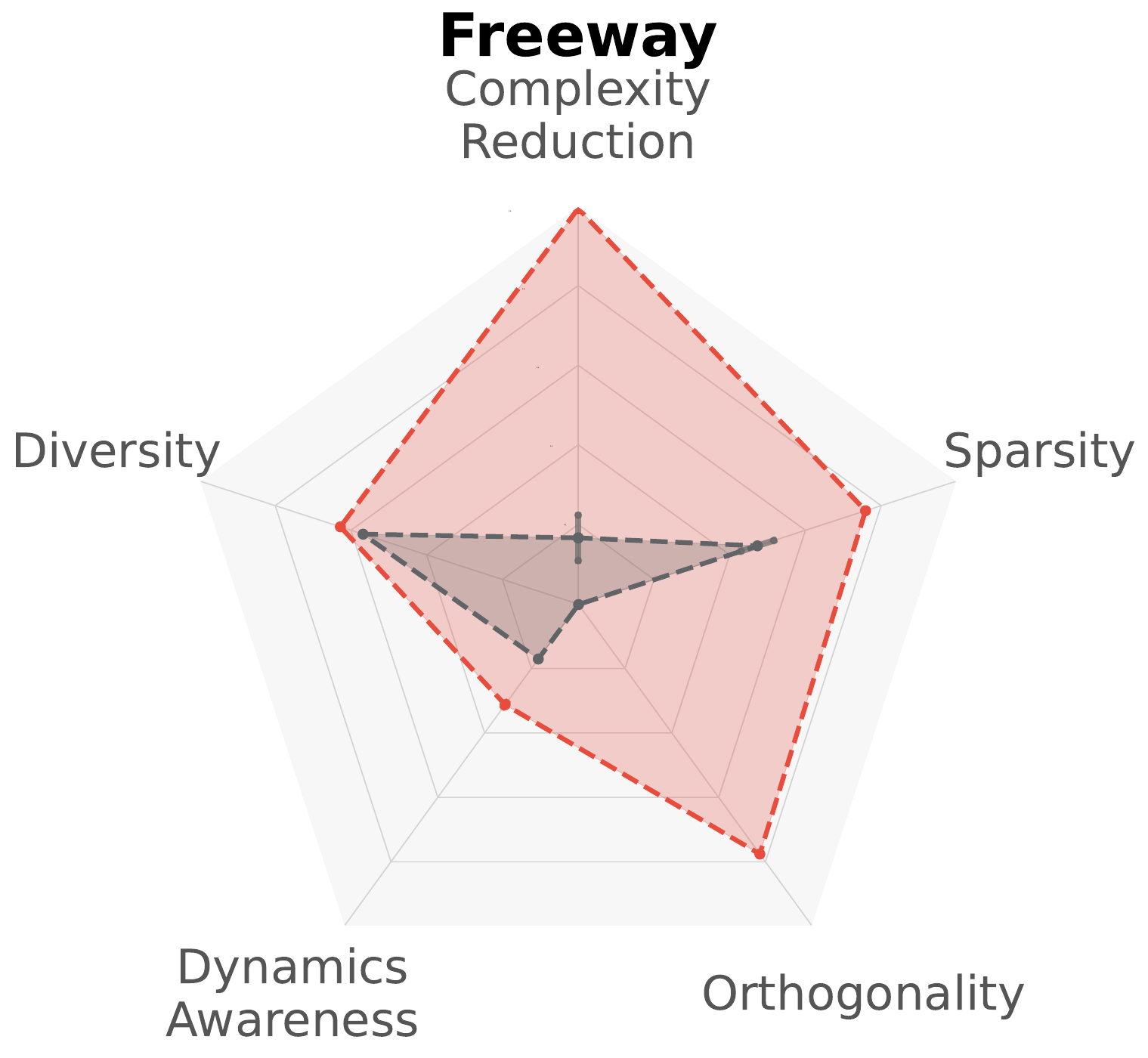}};
		\end{tikzpicture}
	\end{subfigure}   
	\begin{subfigure}[b]{0.32\linewidth}
		\begin{tikzpicture}
		\node (img)  {\includegraphics[width=\linewidth]{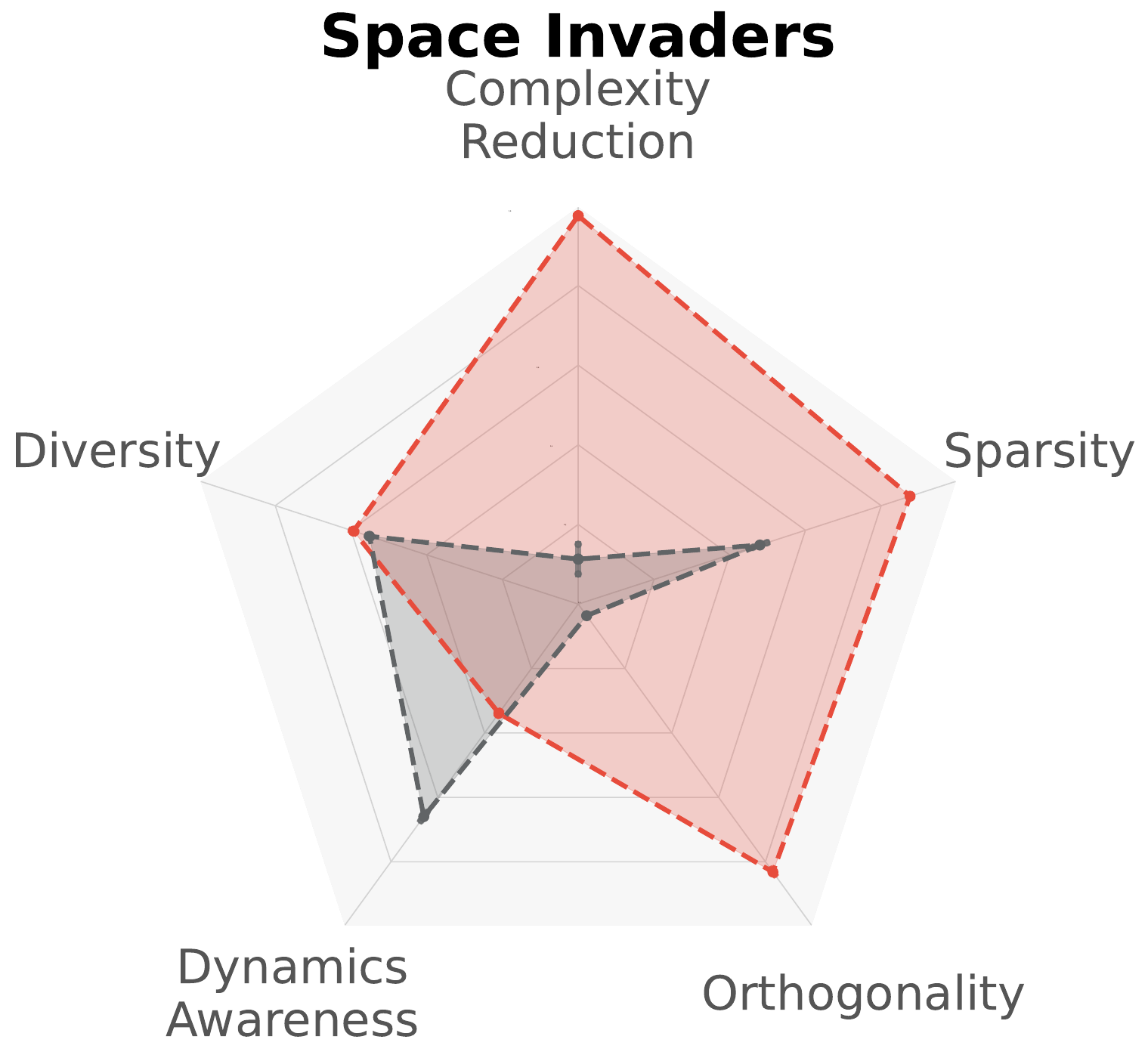}};
		\end{tikzpicture}
	\end{subfigure} 
	\begin{subfigure}[b]{0.32\linewidth}
		\begin{tikzpicture}
		\node (img)  {\includegraphics[width=\linewidth]{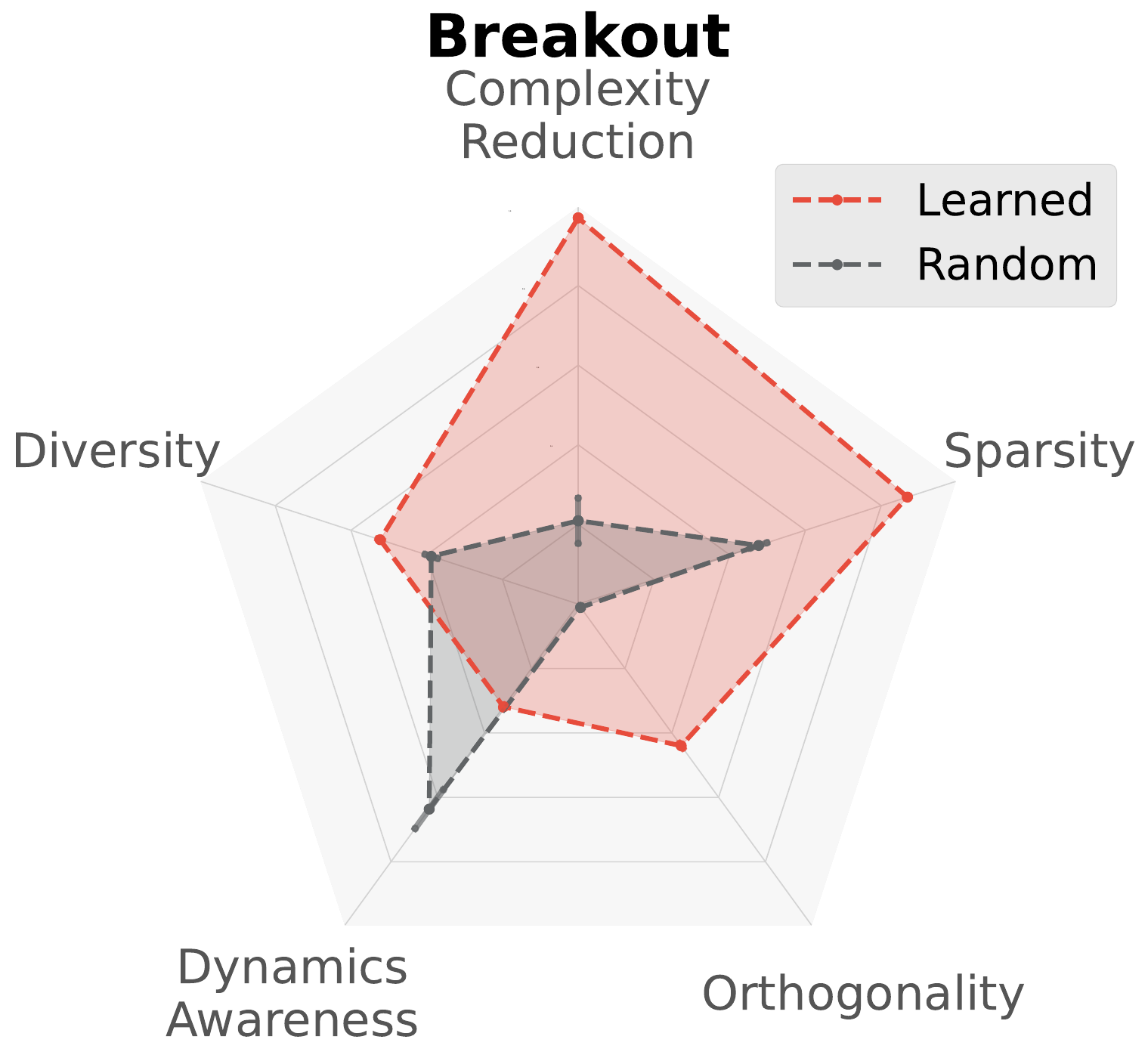}};
		\end{tikzpicture}
	\end{subfigure}  
	\vspace{-.25cm}
	\caption{Properties for the representations learned by Rainbow in Atari. 
		\label{fg:atari_properties}}
\end{figure*}

Next, we consider the properties of the representation produced by Rainbow in the source game mode. We contrast these properties to the representation at initialization, to see how much Rainbow changed the representation properties after training. In the radar plots in Figure \ref{fg:atari_properties}, we see that in general Rainbow increased all the properties, except for dynamics awareness. The most stark change was in complexity reduction, which was significantly increased (see Table~\ref{tb:raw-capacity} in Appendix \ref{app_complexity} for unnormalized complexity--L\_rep--results). This matches our previous results in the maze, where high complexity reduction was key. We also see that Rainbow significantly increases sparsity and orthogonality. Recall that one of our hypotheses for the success of FTA was the fact that it was more naturally able to obtain higher sparsity and orthogonality. We see that diversity only slightly increases, similar to our results under FTA, though different from our previous ReLU network. Both FTA and Rainbow produced more sparse and orthogonal  representations than the ReLU network, which may explain why diversity increased for them both but not for ReLU. 

Finally, dynamics awareness was brought to a medium-low value for all environments (increased in Freeway and significantly decreased in the other two). This outcome is different from what we observed in the maze experiments, where dynamics awareness was significantly increased to a relatively high value. 
One difference is that the random representations for Space Invaders and Breakout start with relatively high dynamics awareness: a random initialization makes nearby points look more similar and more distant points more dissimilar. 
Decreasing the dynamics awareness could happen either due to making temporally distant points have more similar representations or to making representations for successor states very different. 
We can speculate further why this happens.
In the Maze and in Freeway the learned representations exhibited higher dynamics awareness compared to a random representation. In both environments the dynamics are regular throughout training: the location of walls and the movement of cars are independent of the agent's actions. Breakout and Space Invaders are very different. The games start with many objects on the screen and a successful agent learns to destroy all of them.
A well trained representation might produce a smoothly generalizing function with mostly similar states along the trajectory of a well played game. Regardless of the explanation, our result suggests high dynamics awareness is environment specific and perhaps not necessary for good performance.



\section{Conclusions}
The goal of this work is to make progress towards an answer to a classic question: how do the properties of representations---that emerge under standard architectures used in reinforcement learning---relate to the transfer performance? 
We introduced a method of measuring the similarity between training and transfer tasks and designed experiments to assess learned representations. All tasks are similar, in that they involve navigating to locations in the same Maze. Intuitively, transfer should be possible, even to locations that are quite far from the goal in training. 
We found that 1)  ReLU-based representations transferred only to very similar tasks
2) some auxiliary tasks improved transfer of ReLU-based representations, but none facilitated transfer to less similar tasks, 3) the FTA activation significantly improved transfer, suggesting it might be a promising activation to use going forward, and 4) transfer was not possible with a linear value function, even in this seemingly simple environment. 

We extensively and systematically investigated the properties of all of these (good and bad) representations attempting to better understand what causes the improvement in transferability. We defined diversity, complexity reduction, and dynamics awareness, as well as used measures of orthogonality, sparsity and non-interference from the literature. 
In general, interim values for properties were better: representations at the very extremes were never the best. Further, we found that the best representations maintained high complexity reduction, medium-high dynamics awareness, medium diversity and medium orthogonality and sparsity.
These conclusions do not mean representations should have medium orthogonality, for example, but rather representations that emerge under training with auxiliary losses tend to do more poorly if orthogonality is higher or if it is very small (at the extreme), in our transfer setting. 

One message from this work is that it was important to use a simpler environment to develop a systematic methodology and obtain comprehensive results. Even in just this setting, there was a mountain of data to analyze. 
Further, results in this simple environment were already informative and changed our perspective on these representations. A priori, one might have thought that transfer would be very easy in this environment; after all, we are not learning small networks here! Yet, repeatedly we hit roadblocks.
The specific conclusions about network architectures and activations, auxiliary losses, and even properties, may be different in other environments, but the higher-level conclusions about the relevance of these properties, the interactions between components, and the need for a careful methodology to understand these nuances extends. 

We demonstrated the generality of the approach by using the same methodology to analyze the representation of a Rainbow agent that transfers across Atari modes. We found that---similarly to the high-performing representations in the maze environment---the representation had high complexity reduction, higher levels of orthogonality and sparsity and interim diversity. A key difference was that the dynamics awareness was medium-low, rather than medium-high. This difference might indicate that this property is useful for certain environments, like cost-to-goal environments, and not a useful property for representations to have in other environments, like Atari games. 

There are many possible next steps to use this methodology to improve our understanding of representations in reinforcement learning. 
There has been substantial effort to characterize transfer, generalization, and overfitting in deep reinforcement learning, primarily in terms of performance \cite{packer2018assessing,rajeswaran2017towards,cobbe2018quantifying}. A natural next step is to build on this work: repeat their experiments and measure the properties of the learned representations.
Another important next step is to consider correlation between properties and transfer when fine-tuning the representation. In this work, we tested fixed features, to ask if initial learning produced useful features for future learning. It is also useful to ask if initial learning produced a useful starting point or initialization for future learning. We may find that different representation properties are useful in this fine-tuning regime.

\section*{Acknowledgments}

We thank Vlad Mnih for reviewing an earlier version of this paper. This work was generously funded by the CIFAR Canada AI Chair program, Amii and NSERC. 

\bibliographystyle{elsarticle-num} 
\bibliography{refs.bib}


%
%
%


\appendix
\definecolor{random}{rgb}{0.86, 0.08, 0.24}
\definecolor{no-aux}{rgb}{0.55, 0.0, 0.55}
\definecolor{decoder}{rgb}{0.65, 0.16, 0.16}
\definecolor{expert-xy}{rgb}{1.0, 0.89, 0.77}
\definecolor{expert-color}{rgb}{0.72, 0.53, 0.04}
\definecolor{expert-count}{rgb}{0.87, 0.72, 0.53}
\definecolor{pick-red}{rgb}{0.36, 0.54, 0.86}
\definecolor{single-goal}{rgb}{0.36, 0.54, 0.86}
\definecolor{all-goals}{rgb}{0.12, 0.56, 1.0}
\definecolor{next-as}{rgb}{1.0, 0.5, 0.0}
\definecolor{sucgreen}{rgb}{0., 0.5, 0}
\definecolor{reward}{rgb}{0, 0, 0}

\ifx\theorem\undefined
\newtheorem{theorem}{Theorem}
\fi

\ifx\definition\undefined
\newtheorem{definition}{Definition}
\fi

\section{More on Some Representation Properties}
\label{app:moreonreps}

Interestingly, orthogonality in the representations,  as measured in the paper, has a strong relationship with three other properties. Specifically, orthogonality in the representations can be the equivalent to orthogonality between features, which implies that each feature captures distinct information from the input states. Orthogonality can also result in interference, where interference measures how much an update on one state interferes with the updates on other states. Lastly, orthogonal features can be indicators of linearly uncorrelated features.

In this section, we show the aformentioned relationships between orthogonal representations, orthogonal features, interference, and uncorrelated features. To do so, we make the assumption of having a finite set of input-states $\{s_1,s_2, \dots, s_n\}$ represented in a $d$-dimensional space, where the corresponding representations are $\phi(s_i) = [f_1(s_i),f_2(s_i),\dots,f_d(s_i)]^\top$ ($f_l$ being a function to produce the $l$-th feature dimension), and features are $\psi(f_i) = [f_i(s_1), f_i(s_2),\dots, f_i(s_n)]^\top$.




\subsection{Relationship between Orthogonal Representations and Orthogonal Features}\label{app_orth_relationship}

Here, we show that there is an equivalence relationship between orthogonality in the representation and between features. Below we show that $\sum_{i,j}^n (\phi(s_i)^\top\phi(s_j))^2 = \sum_{k,l}^d (\psi(f_k)^\top\psi(f_l))^2$.

\begin{align*}
    \sum_{i,j}^n (\phi(s_i)^\top\phi(s_j))^2 =& \sum_{i,j}^n (\phi(s_i)^\top\phi(s_j)) (\phi(s_i)^\top\phi(s_j)) \\
    =& \sum_{i,j}^n \sum_{k=1}^d f_k(s_i)f_k(s_j) \sum_{l=1}^d f_l(s_i)f_l(s_j) \\
    =& \sum_{i,j}^n \sum_{k,l}^d f_k(s_i)f_l(s_i)f_k(s_j)f_l(s_j) \\
    =& \sum_{k,l}^d \sum_{i,j}^n f_k(s_i)f_l(s_i)f_k(s_j)f_l(s_j) \\
    =& \sum_{k,l}^d \sum_{i=1}^n f_k(s_i)f_l(s_i)\sum_{j=1}^n f_k(s_j)f_l(s_j) \\
    =& \sum_{k,l}^d (\psi(f_k)^\top\psi(f_l)) (\psi(f_k)^\top\psi(f_l)) \\
    =& \sum_{k,l}^d (\psi(f_k)^\top\psi(f_l))^2
\end{align*}

Therefore, when the sample-space is not enumerable, that is $f_i$ is infinite-dimensional, orthogonality of representations may be used as a surrogate for measuring the orthogonality of features.

\subsection{Relationship between Orthogonal Representations and Interference}\label{interf_orth_relationship}

Under linear updating, orthogonal representations would reduce interference. This is because, if two states $s_1$ and $s_2$ have $\phi(s_1)^\top \phi(s_2) = 0$, then performing an update in $s_1$, $\tilde{w} = w + \alpha \delta \phi(s_1)$ has no impact on the prediction in $s_2$: $\phi(s_2)^\top \tilde{w} = \phi(s_2)^\top(w + \alpha \delta \phi(s_1)) = \phi(s_2)^\top w  + \alpha \delta \phi(s_2)^\top\phi(s_1) =  \phi(s_2)^\top w$.

\subsection{Relationship between Orthogonal and Uncorrelated Features}\label{app_orth_correlation}

Here, we show the relationship between orthogonal features and uncorrelated features. Let $\bar \psi(f_i) = [\bar f_i, \bar f_i,\dots, \bar f_i]^\top$, where $\bar f_i = \frac{1}{n} \sum_{j=1}^n [f_i(s_j)]$, denote the expected value of feature $i$ over the set of input-states.
If all features are centered, that is, $\bar f_i = 0$ for all $i$, then it is trivial to see that
\begin{align*}
    \frac{1}{n^2} \sum_{k, l}^d (\psi(f_k) - \bar \psi(f_k))^\top (\psi(f_l) - \bar \psi(f_l)) \\
    = \frac{1}{n^2} \sum_{k,l}^d \psi(f_k)^\top \psi(f_l).
\end{align*}
The LHS is a measure of correlation and the RHS is a measure of orthogonality.





\section{Empirical Details}
\label{app:empiricaldets}

\subsection{Auxiliary Tasks}
\label{app_aux}

\renewcommand{\tabularxcolumn}[1]{m{#1}}

\begin{sidewaystable*}[t!]
    \centering
    \caption{Auxiliary losses used in this paper, defined with respect to transitions in $\mathcal{D}$.}
    \begin{tabularx}{\textwidth}{|@{}>{\centering\arraybackslash\hsize=.12\hsize}m{15cm}|X@{}|}
      \hline
      \centering
      ATC & 
    \begin{equation*}
        \begin{gathered}
        \scalebox{0.8}{
        $\mathcal{L}_{A T C}(\mathcal{D})=-\mathbb{E}_{\mathcal{S}}\left[\log \frac{\exp l_{i,i+k}}{\sum_{s_{j} \in \mathcal{S}} \exp l_{i;j+k}}\right]$, \hspace{0.1cm} 
        \text{where $l_{t;k}$ are logits computed bilinearly such that $l_{i,j+k} = p_i W c_{j+k}$ }}
    \end{gathered} 
    \end{equation*}
    \begin{equation*}
    \text{and} \hspace{0.1cm}  p_t = F_{\theta_C}(\auxfunc(\repfunc(\text{AUG}(s_t))))+ \auxfunc(\repfunc(\text{AUG}(s_t))), \hspace{0.1cm}
    c_{t+k} = F_{\breve{\theta}_A}(\phi_{\breve{\theta}_R}(\text{AUG}(s_{t+k})))    
    \end{equation*} \\ \hline
      IR & \begin{equation*} \mathcal{L}_{\text{IR}}\left(\mathcal{D}\right)=\underset{s \sim \mathcal{D}}{\mathbb{E}}\left[\| \auxfunc(\repfunc(s))-s\|_{2}^{2}\right] \end{equation*} \\ [-1em] \hline
      NAS & 
      \begin{equation*}
      \begin{gathered}
      \scalebox{0.7}{
      $\mathcal{L}_{\text{NAS}}\left(\mathcal{D}\right)  = \underset{\begin{subarray}{c}
        (s_t, a_t, s_{t+1}) \sim \mathcal{D} \\
        (s_{k \neq t},  a_{k \neq t})\sim \mathcal{D}
        \end{subarray}}{\mathbb{E}}
        \Big[\| \auxfunc(\repfunc(s_t), a_t) - (\repfunc(s_{t+1}) - \repfunc(s_{t}))\|_{2}^{2} +  \max \left(0, 1 - \|\auxfunc(\repfunc(s_k), s_k) - (\repfunc(s_{t+1}) - \repfunc(s_{t}))\|_{2}^{2}\right)\Big] $}
    \end{gathered} 
    \end{equation*} \\ [-1em]\hline
    Reward & \begin{equation*}\begin{gathered}
    \mathcal{L}_{\text{Reward}}\left(\mathcal{D}\right)=\underset{(s_t, a_t, r_{t+1}) \sim \mathcal{D}}{\mathbb{E}}\left[\| \auxfunc(\repfunc(s_t), a_t)-r_{t+1}\|_{2}^{2}\right]\end{gathered} \end{equation*}\\[-1em]\hline
      SF &  \begin{equation*} \mathcal{L}_{\text{SF}}\left(\mathcal{D}\right) = \underset{(s_t, a_t, s_{t+1}, a_{t+1})  \sim \mathcal{D}}{\mathbb{E}}\left[\| \auxfunc(\repfunc(s_t), a_{t}) - (\repfunc(s_t) + \gamma \auxfunctarget(\repfunctarget(s_{t+1}), a_{t+1}))\|_{2}^{2}\right] \end{equation*} \\ [-1em] \hline
      VirtualVF & \begin{equation*} \mathcal{L}_{\text{VirtualVF}}\left(\mathcal{D}\right) = \sum_{g \in \mathcal{G}} \underset{(s_t, a_t, r_{t+1}, s_{t+1}, a_{t+1})  \sim \mathcal{D}}{\mathbb{E}}\left[\| F^{g}_{\theta_A}(\repfunc(s_t), a_{t}) - (r^{g}_{t} + \gamma F^{g}_{\hat{\theta}_A}(\repfunctarget(s_{t+1}), a_{t+1})\|_{2}^{2}\right]  \end{equation*}  \\ [-1em]\hline
      XY & 
      \begin{equation*} 
      \mathcal{L}_{\text{XY}}\left(\mathcal{D}\right)=\underset{(s_t, x_t, y_t) \sim \mathcal{D}}{\mathbb{E}}\left[\| \auxfunc(\repfunc(s_t)) - (x_t, y_t)\|_{2}^{2}\right] 
      \end{equation*} 
      \small $\mathcal{G}$ is a set of goal locations. The reward $r_g$ and representation function $\phi^{g}_{\theta_A}$ are associated with the goal $g$, sampled from this set. \\ \hline
    \end{tabularx}
    \label{table:aux-loss}
\end{sidewaystable*}

In this section, we provide a more detailed explanation of each of the seven auxiliary tasks used for helping with representation learning. All the losses use the same samples taken from the replay buffer to update the value function, except the ATC loss. The detailed formulas for the auxiliary losses are presented in Table \ref{table:aux-loss}. 

\textbf{Augmented Temporal Contrast (ATC) Loss:} This contrastive loss encourages the network to learn similar representations for an input state, $s_{t}$, with one from a pre-determined, near-future time step input state, $s_{t+k}$, where $k=3$. In contrast to other losses that we have employed in this work, this loss uses more than a single auxiliary head to compute itself. To do so, it first applies a data augmentation technique called random shift with a probability of $0.1$ and padding of $4$ to both of these input states. Then, it feeds the augmented version of $s_{t}$, $\text{AUG}(s_{t})$, through a set of networks to compute $p_t$, where $F_{\theta_A}$ is a linear mapping of representation into an embedding space with a size of $32$, and $F_{\theta_C}$ is a single layer neural network with a hidden layer size of $64$, and an output size of $32$. Then, $\text{AUG}(s_{t+k})$ is fed into a momentum encoder of $\repfunc$ ($\phi_{\breve{\theta}_R}$) and $F_{\theta_A}$ ($F_{\breve{\theta}_A}$) to compute $c_{t+k}$. The output of these networks, $p_t$ and $c_{t+k}$, are combined with each other through a $32\times32$ matrix called $W$ to compute logits, $l{i,j+k}$. At last, these logits are used to compute the InfoNCE loss, $\mathcal{L}_{A T C}$. In contrast to the original paper that uses a complex learning-rate scheduling technique, we implement this loss in its simplest form by using a fixed learning-rate. We sweep through values of $[0.003, 0.001, 0.0003, 0.0001, 0.00003, 0.00001]$ of this learning-rate. We update the momentum encoder in every step using a $\tau$ of $0.01$. The batch size is the same as the batch size used for computing the value function loss, and the weight of this auxiliary loss is set to $1$.

\textbf{Input Reconstruction (IR):} This auxiliary task reconstructs the input image from the representation through a deconvolutional network. On the auxiliary head, the representation was firstly projected to a hidden layer with $1024$ nodes, then sent to a two layer deconvolutional neural network, with $4$ kernels, $32$ and $3$ channels, $2$ and $1$ strides, $2$ and $1$ pads on $2$ layers separately. The output had the same size as the input image (i.e., $15\times15\times3$). The weight of the auxiliary loss was set to $0.0001$.

\textbf{Next Agent State Prediction (NAS):} This task motivates the representation of the current state to minimize its prediction error on the next state representation. To do so, it uses a contrastive loss that minimizes the prediction error on the next state while maximizes the prediction error on the rest of the states. For this task, the auxiliary head, $F_{\theta_A}$, consists of two fully connected layers with $64$ neurons on each. The weight of this auxiliary loss was $0.001$.


\textbf{Successor Feature Prediction (SF):}
Successor feature prediction task was similar to next-agent-state prediction, though the target was constructed by bootstrapping. Given the transition on latent space $<\phi_t, a_t, \phi_{t+1}, a_{t+1}>$, the auxiliary head learned to minimize the difference between the prediction $F_{\theta_A}(\phi_{t}, a_t)$ and the target $(1-\lambda)\phi_{t+1} + \lambda F_{\theta_A}(\phi_{t+1}, a_{t+1})$, where $\lambda$ is set to $0.99$. To satisfy the property of successor features, we added an extra head to predict the reward linearly from the representation $\phi_t$. This head for predicting the successor feature used the same neural network architecture as NAS. The weight was set to~$1$.

\textbf{Reward Prediction (Reward):}
Another auxiliary prediction task was to predict the reward independently given the representation $\phi_t$. This auxiliary task used the same non-linear transformation structure as SF and the weight is set to~$1$.

\textbf{Expert Target Prediction (XY):} 
The last prediction task was expert-designed targets prediction. The agent was asked to predict its current position given the image. Since predicting the position was considered a regression task, we used MSE loss with the same network structure as Reward. We set a low weight of $0.0001$ for this auxiliary task.


\textbf{Virtual Value Function Learning (VirtualVF):} 
This auxiliary task learns a different value function on the auxiliary head. 
There were $2$ settings in Maze---learning a value function assuming the goal is on grid $[7, 7]$, and learning $5$ value functions when the goals are on grid $[0, 0]$, $[0, 14]$, $[14, 0]$, $[14, 14]$, $[7, 7]$ separately.
The weight of this auxiliary task remained $1$ but the discount rate on the auxiliary head was set to be lower, $0.9$, so that the agent can focus on the main task. The auxiliary head learned this task with the same network structure as XY.

\subsection{General Hyperparameter Setting}\label{app:representation_learning}

We used the same neural network architecture across representations learned with the various loss functions, for each domain. All hidden layers are initialized with Xavier.
Table \ref{tb:rep_size} shows the number of nodes on the representation function's last hidden layer, and the number of features.

\begin{table}[!t]
	\renewcommand{\arraystretch}{1.3}
	\caption{Representation size settings}
	\label{tb:rep_size}
	\centering
	\begin{tabular}{c c c c}
		Label & Activation & Last Hidden Layer Nodes & Features \\ 
		\hline
		ReLU  & ReLU & 32 & 32 \\
		ReLU(L)  & ReLU & 640 & 640 \\
		FTA  & FTA & 32 & 640 \\
	\end{tabular}
\end{table}

During training, the inputs are normalized to be in the range $[-1, 1]$. We use Adam optimizer to update weights, and we used the mean-squared error as the loss. The batch size is set to be $32$. The buffer has length $10,000$. The input image is normalized. 
For the representation function, we use a two layer convolutional network with kernel size of $4$, stride of $1$, padding of $1$, and $32$ channels for the first layer; kernel size of $4$, stride of $2$, padding of $2$, and $16$ channels on the second layer.
A target network was used with the synchronization frequency set to $64$. The buffer's memory size was $100,000$ and there were $32$ samples randomly chosen at each step. The agent learns for $300,000$ steps with $\epsilon$-greedy policy. 
During transfer learning, all agents, including baselines, learned for $100,000$ steps only.

As for the FTA setting, we use $20$ bins with the higher and lower bounds equal to $2$ and $-2$. We tested $\eta$ of $0.2, 0.4, 0.6$, and $0.8$ for the no auxiliary task agent, and we fixed $\eta=0.2$ for agents trained with auxiliary task.

The learning rate was swept for every representation learning architecture and control task. The best setting was picked according to the averaged performance over $5$ runs. Each run uses a different random seed.
We tested the learning rate in $[0.001, 0.0003, 0.0001, 0.00003, 0.00001]$ during the representations training step and for learning from scratch baseline agents. During transfer, we tested $[0.01, 0.003, 0.001, 0.0003, 0.0001]$ for representations with $32$ features and $[0.001, 0.0003, 0.0001, 0.00003, 0.00001]$ for all representations with $640$ features, as there were more weights to train, which required more careful learning. 

In the non-linear value function case and transfer tasks with linear value function, we use a fixed $\epsilon=0.1$. However, in representation learning in the linear case, it turned out to be harder for the agent to converge when keeping other settings as the same as in the non-linear value function. Thus, we provided a better exploration in the early learning stage to speed up learning by decreasing $\epsilon$, which decreases from $1$ to $0.1$ in the first $100,000$ steps.

\subsection{Using SFs to Measure Task Similarity} \label{sec:task_similarity} 
When checking how the difficulty of transfer affects the transfer performance, we consider the similarity between each transfer task to the original task. When a transfer task is similar to the task in which the representation is trained, we consider the transfer to be easier.

We measure the similarity between tasks according to the successor representations, $\psi$. Since successor representations encode the trajectories of the agent, the difference between successor representations generated by optimal policy can reflect the difference between optimal policies that the agent learns in different tasks. If the optimal policies of two tasks turn out to be dissimilar, the similarity between these tasks is considered low.

To compute a highly accurate estimate of successor features, we solve the maze by using a simple tabular algorithm. To do so, we define the state as the cell in the maze that is occupied by the agent. Doing so results in having $173$ states in total. Taking this into account, we use value iteration to generate an optimal policy, then calculate the successor representation of each state based on the optimal policy. 


The successor representations of all states in the same task are considered. The successor representation of each task, $\Psi$, is obtained by concatenating all successor features in the same task. The similarity is defined as the dot product between $\Psi$'s in the transfer and the original tasks.
We choose the dot product to keep both angle and magnitude information between concatenated successor representations. A higher dot product value means the transfer task is more similar to the original task and vice versa. 
\begin{align*}
\psi(s, task_x) &= \mathbb{E}_{\pi^*_{task_x}} \left[ \sum^T_{t=0} \gamma^t f(S_{t}) | S_0=s\right] \\
\Psi_{task_x} &= \left[\psi(s_0) \; \psi(s_1) \; \cdots \; \psi(s_{|\States|}) \right] \\
similarity(task_x, task_y) &= \Psi_{task_x} \cdot \Psi_{task_y}
\end{align*}

\begin{figure}[!t]
	\centering	
	\begin{subfigure}[b]{0.5\linewidth}
		\includegraphics[width=\linewidth]{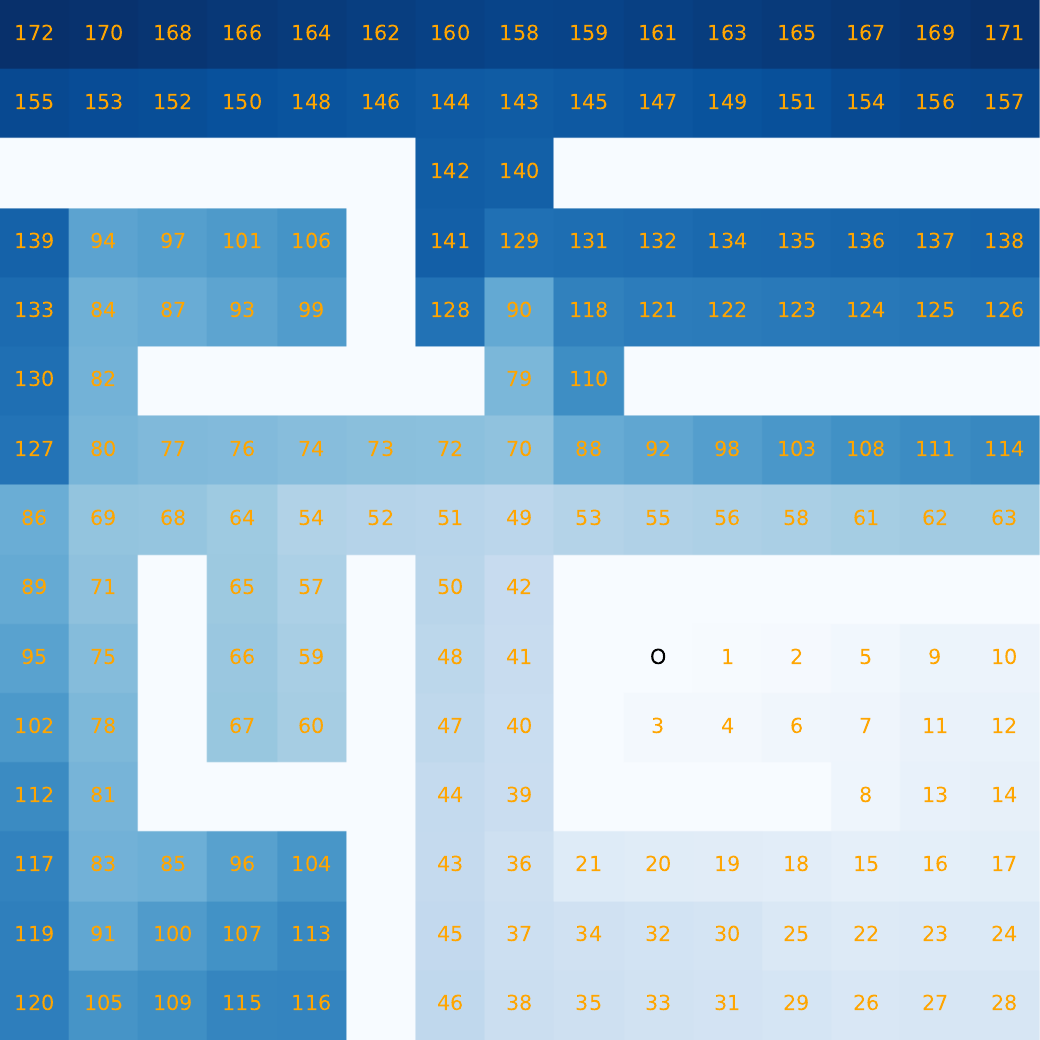}
	\end{subfigure}	
	\caption{Transfer tasks are ranked by the similarity between successor features of each task. This figure shows the similarity ranking of different tasks compared to the source task, where the source task is marked by O. Each number in the cell indicates the similarity rank of the task when the goal is moved to that specific position. 
	}
	\label{fg:goal_ranks}
\end{figure}

Interestingly, the goal states that are more distant from each other become more dissimilar by computing the similarity this way. This is more clear when we take a look at the similarity rankings of the goal states, as depicted in Figure \ref{fg:goal_ranks}. As shown in Figure \ref{fg:auc_chosen}, the representations have a hard time transferring to higher-ranking goals, so there is a clear connection between the ranks of the transfer tasks and the transfer performance. These findings support the use of this approach for calculating task similarity and ranking.

\begin{figure}[htb!]
    \centering  
    \begin{subfigure}[b]{0.15\linewidth}
        \begin{tikzpicture}
        \node[rotate=0, align=center,font=\color{black}] {\footnotesize Return\\ \footnotesize per\\ \footnotesize Episode\vspace{0.7cm}};
        \end{tikzpicture}
    \end{subfigure}
    \begin{subfigure}[b]{0.5\linewidth}
        \begin{tikzpicture}
        \node (img)  {\includegraphics[width=\linewidth]{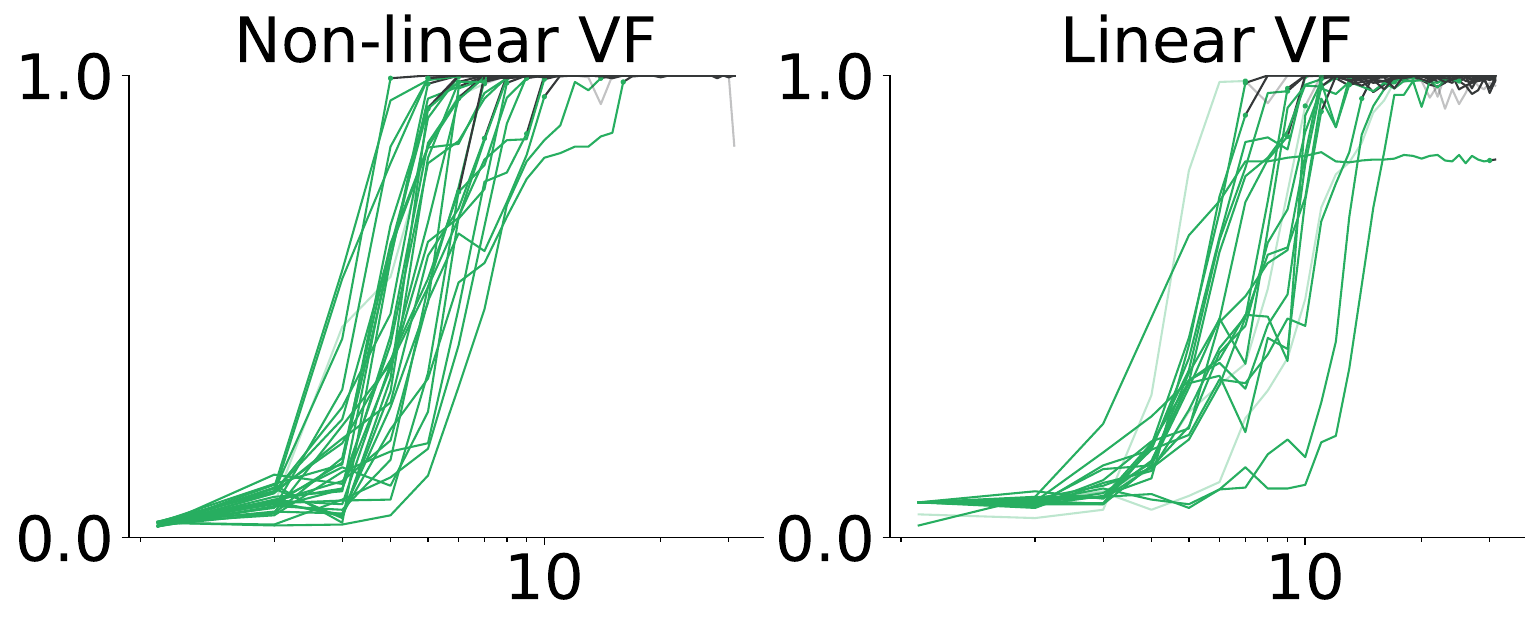}};
        \end{tikzpicture}
    \end{subfigure} 
    
    \begin{subfigure}[b]{\linewidth}
        \centering
        \begin{tikzpicture}
        \node[align=center,font=\color{black}] {\footnotesize Time Steps ($10^4$, logscale)};
        \end{tikzpicture}
    \end{subfigure}
    
    \caption{Return converges in the representation training step.
        The plot shows the averaged return over the most recent 100 episodes at each checkpoint. The x-axis is the number of time steps and the y-axis is the average return.
        Each curve represents one agent specification (activation and auxiliary task pair). As our main focus is not to compare the learning efficiency during the representation learning step, and the difference between learning curves is not large, we only show the general trend by plotting every curve with the same color. The curve changes color to black, at the time point where we took the representation and fixed it.
    }
    \label{fg:rep_training}
\end{figure}
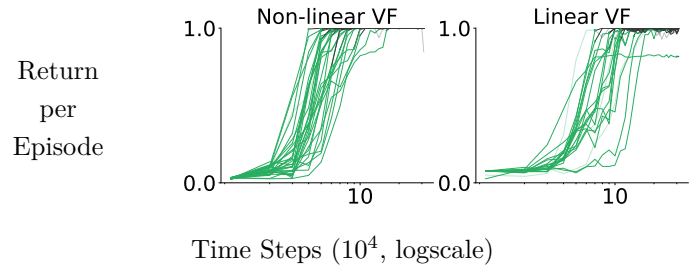

\section{Additional Results} \label{apdx:exp_results}
\subsection{Representation Training} \label{apdx:rep_training}

We show the learning curve of all representation learning architectures in Figure \ref{fg:rep_training}, to show that the early-saved representations have converged when they are saved.

\subsection{Larger ReLU Transfer} \label{apdx:relu_l}
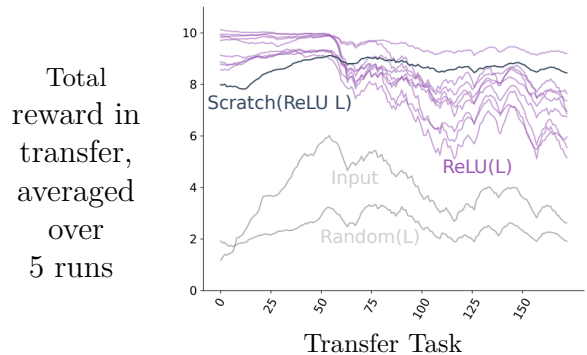
\begin{figure}[!htbp]
	\centering
	
	\begin{subfigure}[b]{0.15\linewidth}
		\begin{tikzpicture}
		\node[rotate=0, align=center,font=\color{black}] {\footnotesize Total\\reward in\\transfer,\\averaged\\over\\5 runs \vspace{1cm}};
		\end{tikzpicture}
	\end{subfigure}
	\begin{tikzpicture}
	\node (img)  {\includegraphics[width=0.4\linewidth]{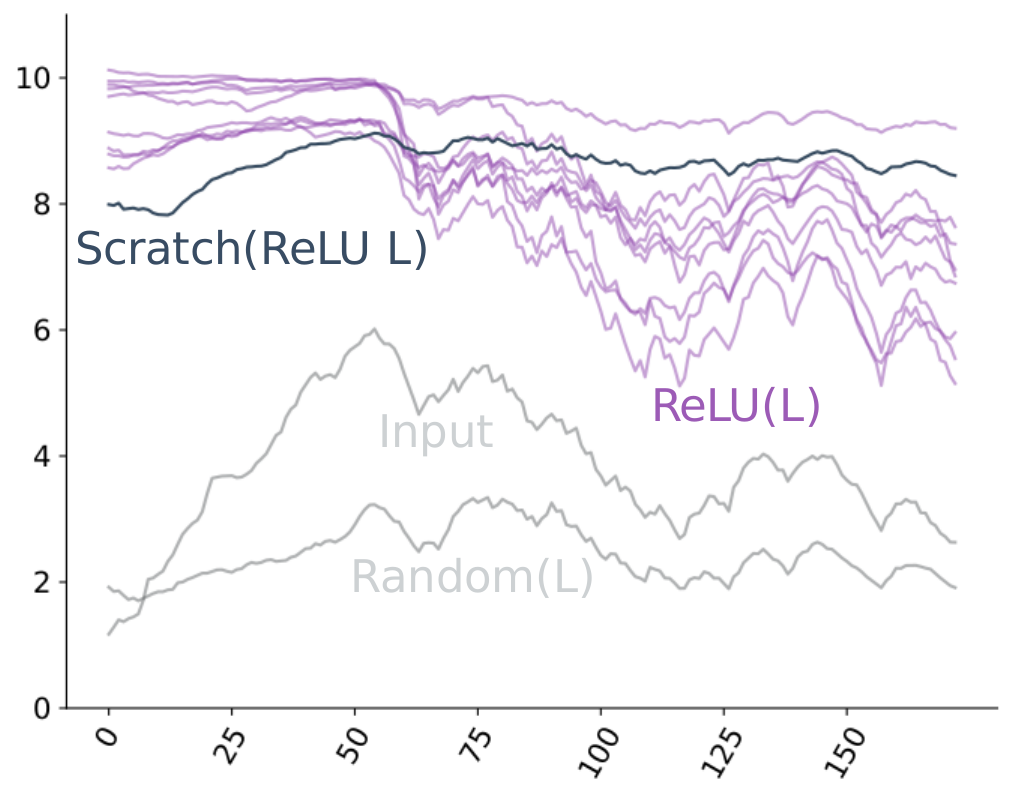}};
	\node[below=of img, node distance=0cm, yshift=1.2cm,font=\color{black}] {\footnotesize Transfer Task};
	
	\end{tikzpicture}
	\caption{
		Transfer performance of larger ReLU representations (45 in total) on 173 transfer tasks. The tasks on the x-axis are arranged by similarity to training tasks: on the left (small x-values) being most similar and on the right (large x-values) being most dissimilar. The black line shows the performance when learning in each transfer task from scratch, with the same representation size as ReLU(L). Lines completely above the black line indicate a representation yielded successful transfer in all tasks. Lines that start above the black line but fall below it as we move left to right indicate a representation that transfers to similar tasks but not dissimilar tasks.
	}
	\label{fg:auc_relul}
\end{figure}
We show the transfer performance of ReLU(L) in Figure \ref{fg:auc_relul}. The ReLU(L) setting stays between ReLU and FTA representations: ReLU(L) keeps the same activation function as ReLU representation, but increases the size of the representation layer to 640, which is the same as the size of FTA representations. Therefore, it maintains the same value function capacity as the FTA representations. The pattern in the transfer performance of ReLU(L) is similar to ReLU (Figure \ref{fg:auc_chosen}). As the transfer tasks become dissimilar, the transfer performance drops below the Scratch agent. In general, when considering the total reward obtained by the agent, the performance of ReLU(L) is better than ReLU and worse than FTA.

\subsection{Transfer with Different Architectures}\label{app_ci}
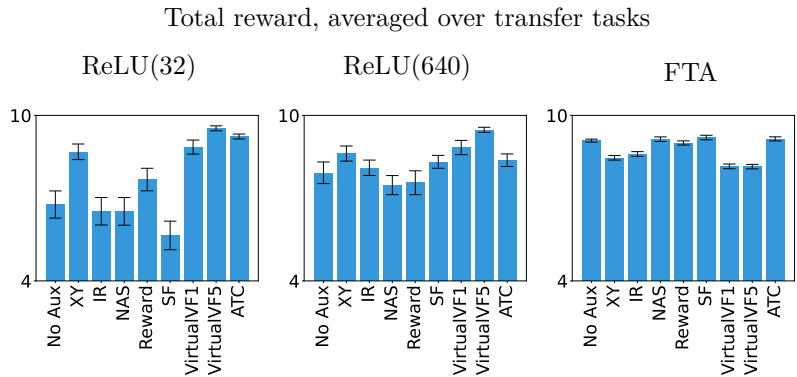
\begin{figure}[!t]
	\centering
	\begin{tikzpicture}	
	\node[rotate=0, align=center, font=\color{black}] {\qquad \quad \footnotesize Total reward, averaged over transfer tasks};
	\end{tikzpicture}
	\begin{subfigure}[b]{0.05\linewidth}
		\begin{tikzpicture}		
		\node[rotate=90, align=center,font=\color{black}] {};
		\end{tikzpicture}
	\end{subfigure}

	\begin{subfigure}[b]{0.25\linewidth}
		\centering
		\begin{tikzpicture}
		\node[align=center,font=\color{black}] {\quad \footnotesize ReLU(32)};
		\end{tikzpicture}
	\end{subfigure}
	\begin{subfigure}[b]{0.25\linewidth}
		\centering
		\begin{tikzpicture}
		\node[align=center,font=\color{black}] {\quad \footnotesize ReLU(640)};
		\end{tikzpicture}
	\end{subfigure}
	\begin{subfigure}[b]{0.25\linewidth}
		\centering
		\begin{tikzpicture}
		\node[align=center,font=\color{black}] {\qquad \footnotesize FTA};
		\end{tikzpicture}
	\end{subfigure}
	
	\begin{subfigure}[b]{0.25\linewidth}
		\begin{tikzpicture}
		\node (img)  {\includegraphics[width=\linewidth]{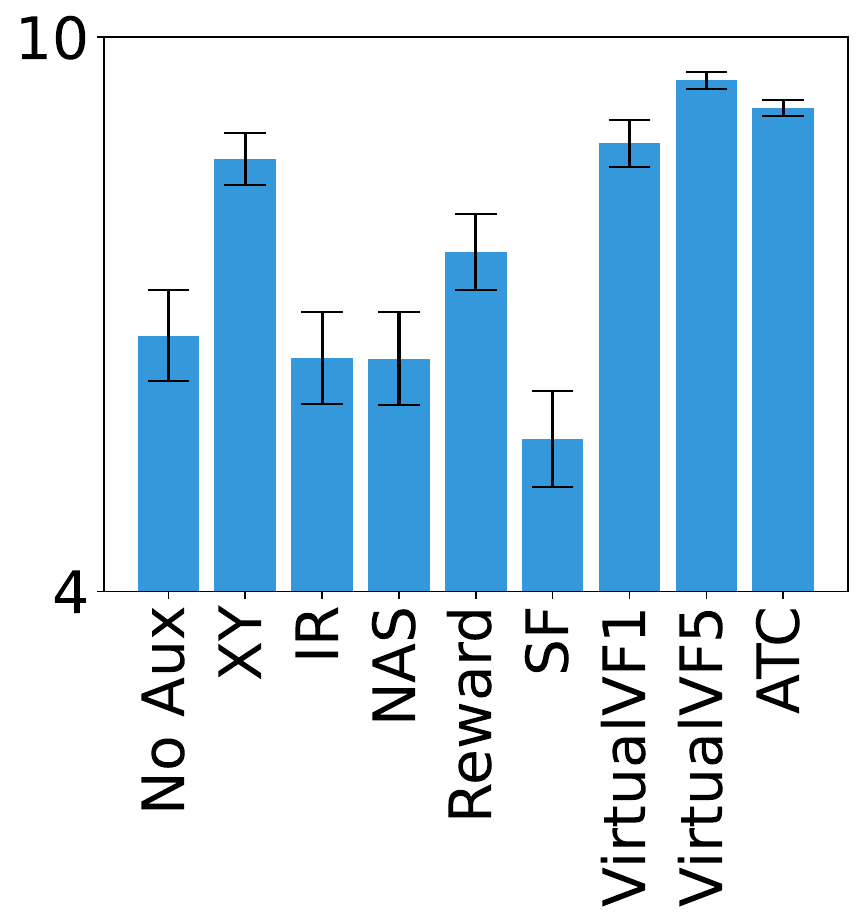}};
		\end{tikzpicture}
	\end{subfigure}	
	\begin{subfigure}[b]{0.25\linewidth}
		\begin{tikzpicture}
		\node (img)  {\includegraphics[width=\linewidth]{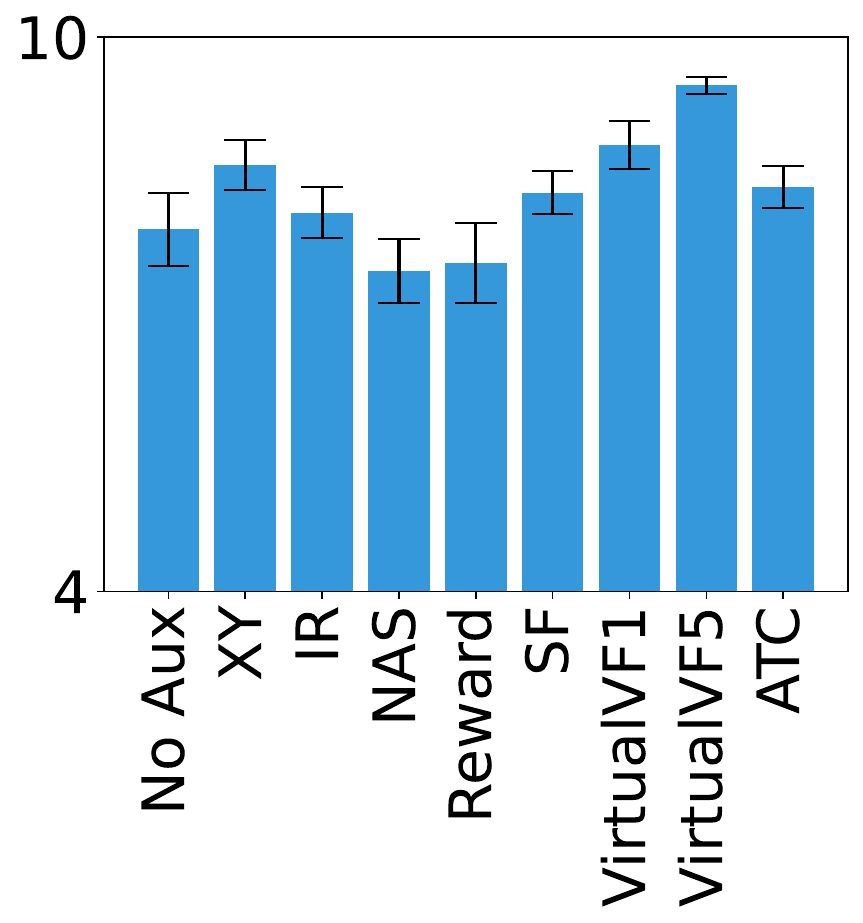}};
		\end{tikzpicture}
	\end{subfigure}
	\begin{subfigure}[b]{0.25\linewidth}
		\begin{tikzpicture}
		\node (img)  {\includegraphics[width=\linewidth]{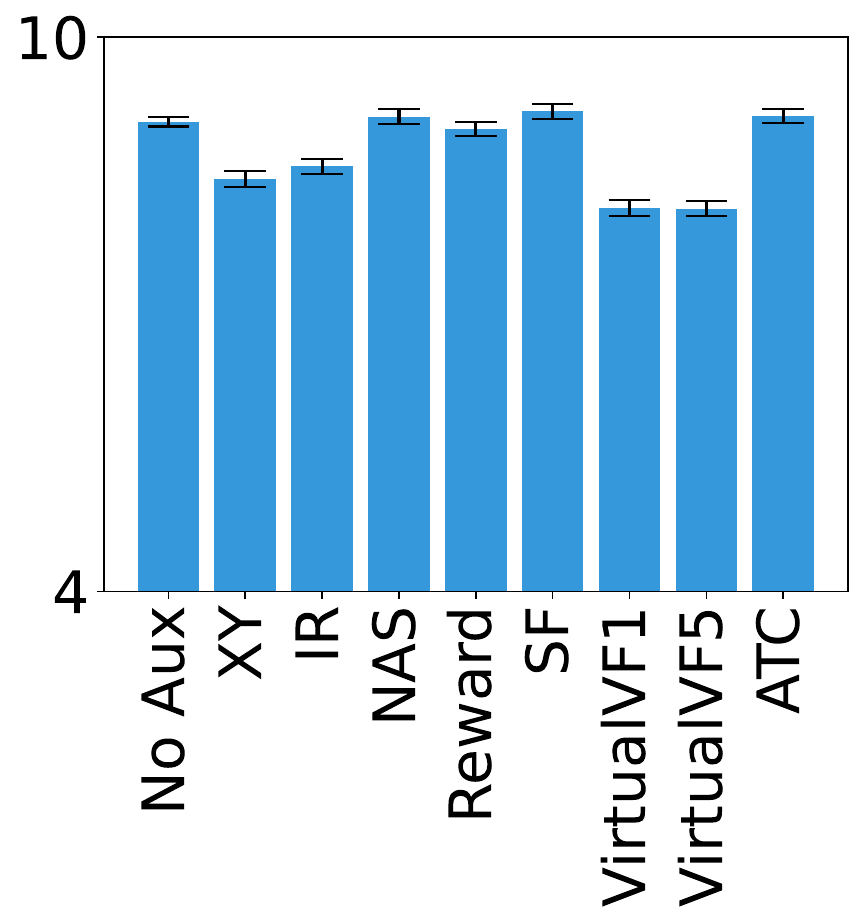}};
		\end{tikzpicture}
	\end{subfigure}
	
	\caption{Transfer performance depends on the activation function, representation size, and auxiliary tasks. This plot presents the same data as Figure \ref{fg:atc_ac_activation}, but the error bar shows a 95\% confidence interval. The bar shows the mean value over 5 seeds $\times$ 173 transfer tasks.}
	\label{fg:atc_ac_activation_95CI}
\end{figure}

Figure \ref{fg:atc_ac_activation_95CI} shows the 95\% confidence interval of transfer performance with different representation sizes, activation functions, and different auxiliary tasks, with a non-linear value function.

\subsection{Relationship between Properties} \label{apdx:between_prop}
\begin{figure}[!htbp]
	\centering	
	\begin{subfigure}[b]{0.24\linewidth}
		\begin{tikzpicture}
		\node (img)  {\includegraphics[width=\linewidth]{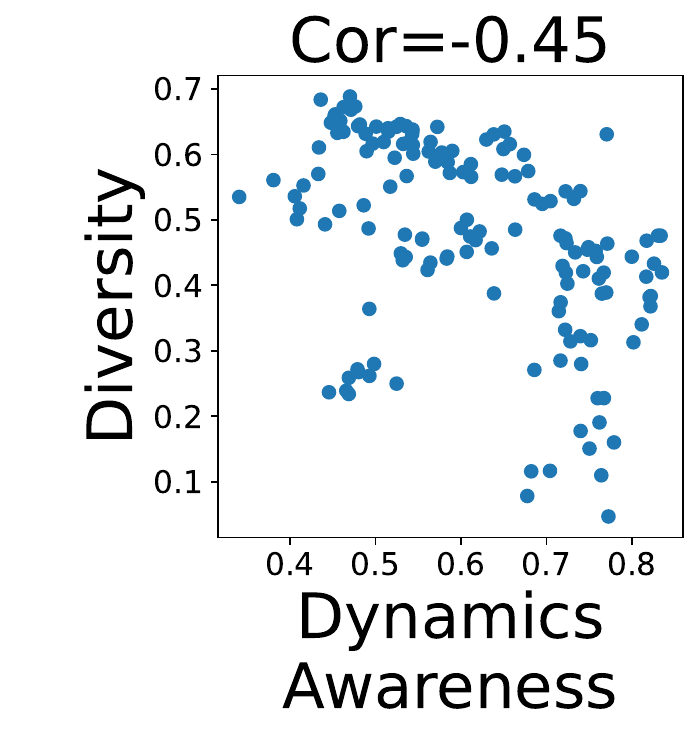}};
		\end{tikzpicture}
	\end{subfigure}
	\begin{subfigure}[b]{0.24\linewidth}
		\begin{tikzpicture}
		\node (img)  {\includegraphics[width=\linewidth]{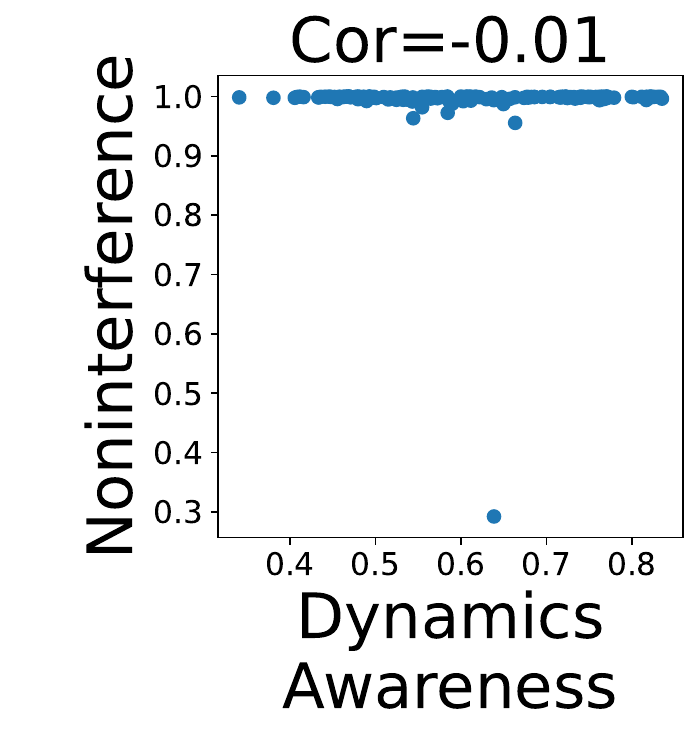}};
		\end{tikzpicture}
	\end{subfigure}
	\begin{subfigure}[b]{0.24\linewidth}
		\begin{tikzpicture}
		\node (img)  {\includegraphics[width=\linewidth]{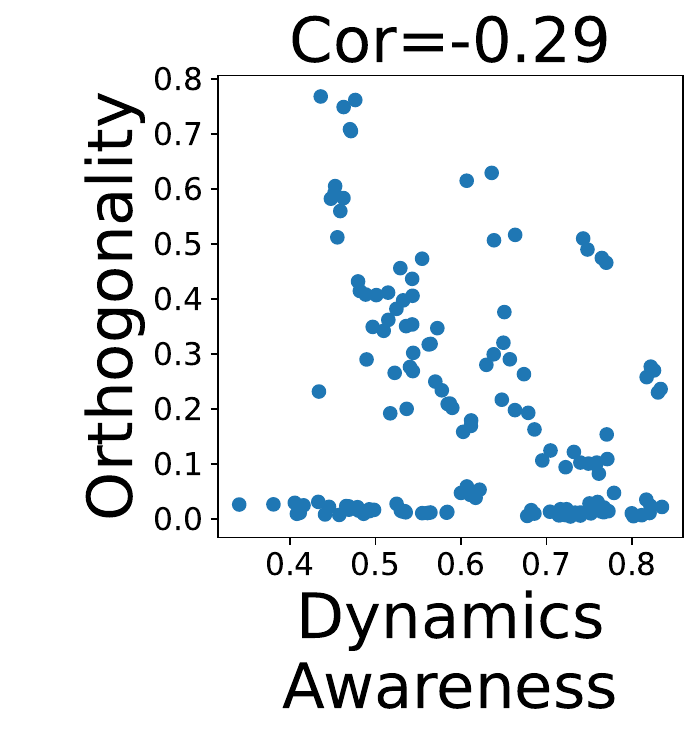}};
		\end{tikzpicture}
	\end{subfigure}
	\begin{subfigure}[b]{0.24\linewidth}
		\begin{tikzpicture}
		\node (img)  {\includegraphics[width=\linewidth]{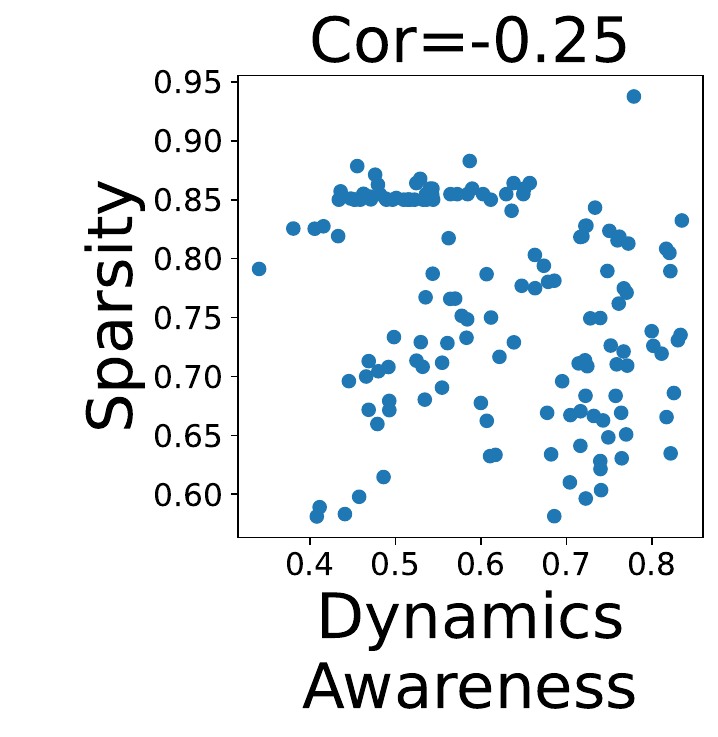}};
		\end{tikzpicture}
	\end{subfigure}

	\begin{subfigure}[b]{0.24\linewidth}
		\begin{tikzpicture}
		\node (img)  {\includegraphics[width=\linewidth]{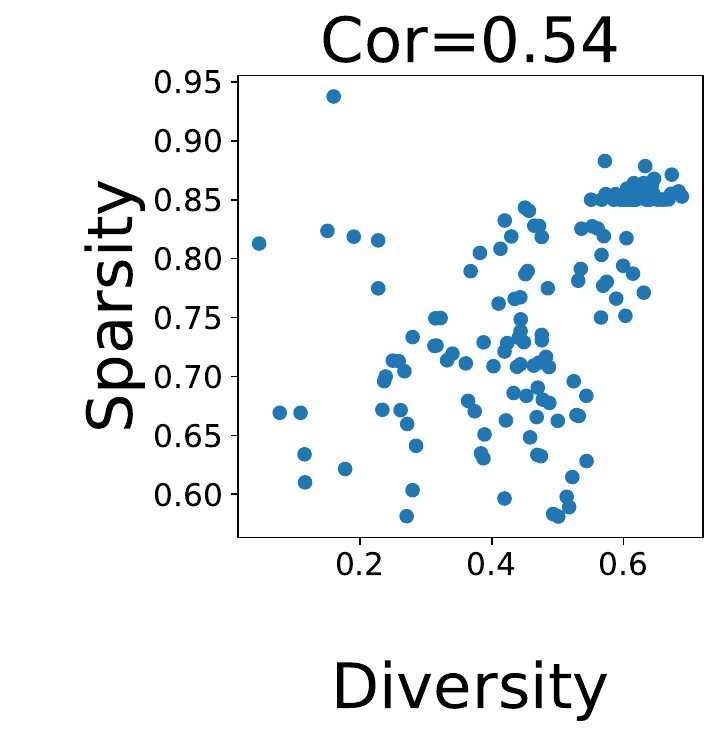}};
		\end{tikzpicture}
	\end{subfigure}
	\begin{subfigure}[b]{0.24\linewidth}
		\begin{tikzpicture}
		\node (img)  {\includegraphics[width=\linewidth]{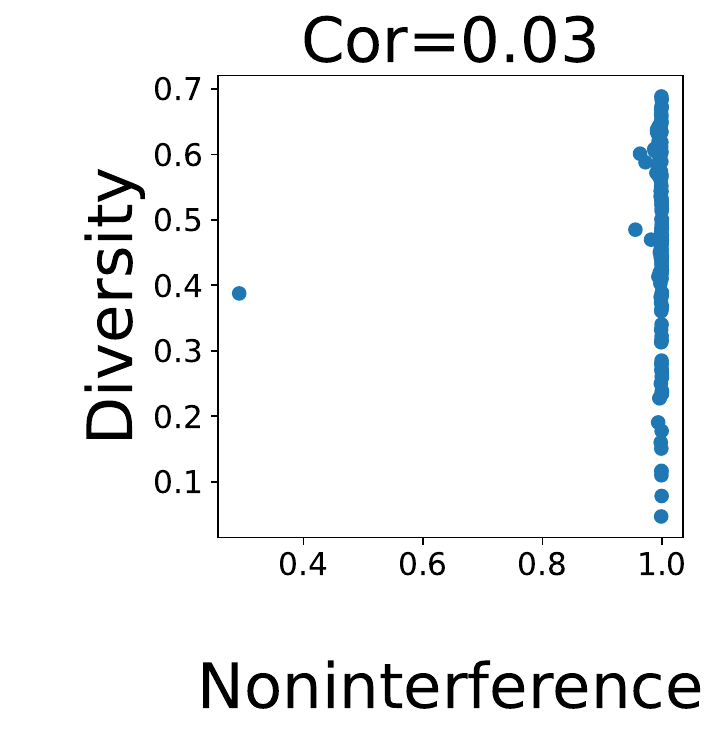}};
		\end{tikzpicture}
	\end{subfigure}
	\begin{subfigure}[b]{0.24\linewidth}
		\begin{tikzpicture}
		\node (img)  {\includegraphics[width=\linewidth]{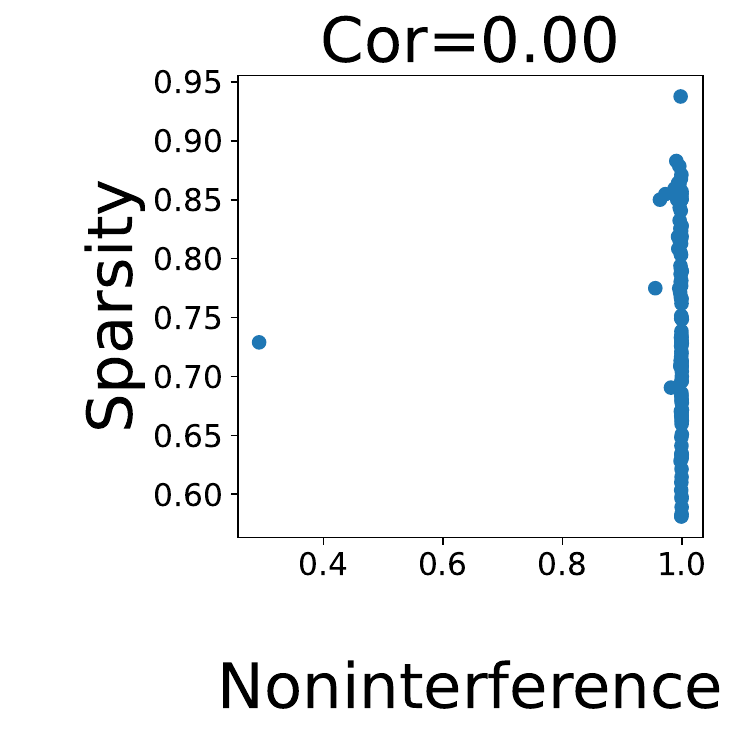}};
		\end{tikzpicture}
	\end{subfigure}
	\begin{subfigure}[b]{0.24\linewidth}
		\begin{tikzpicture}
		\node (img)  {\includegraphics[width=\linewidth]{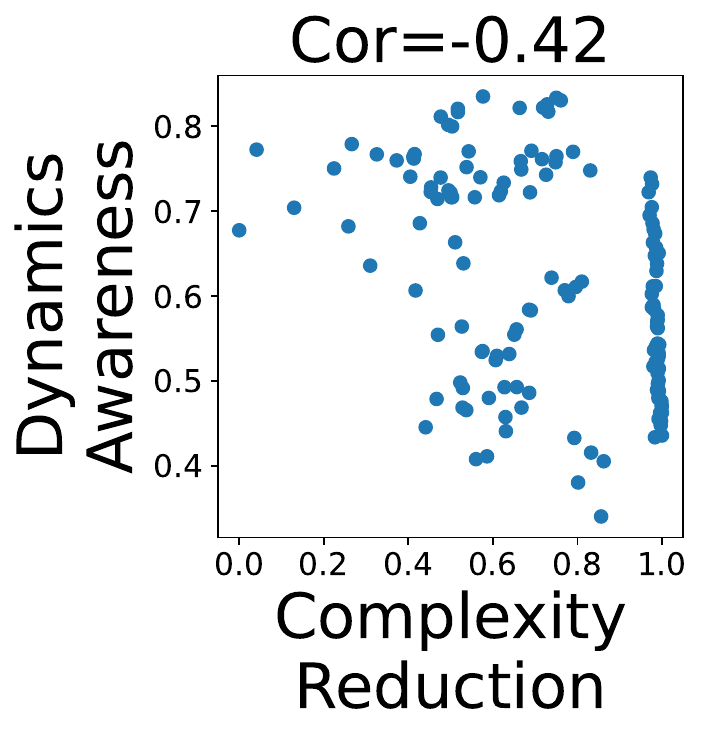}};
		\end{tikzpicture}
	\end{subfigure}

	\begin{subfigure}[b]{0.24\linewidth}
		\begin{tikzpicture}
		\node (img)  {\includegraphics[width=\linewidth]{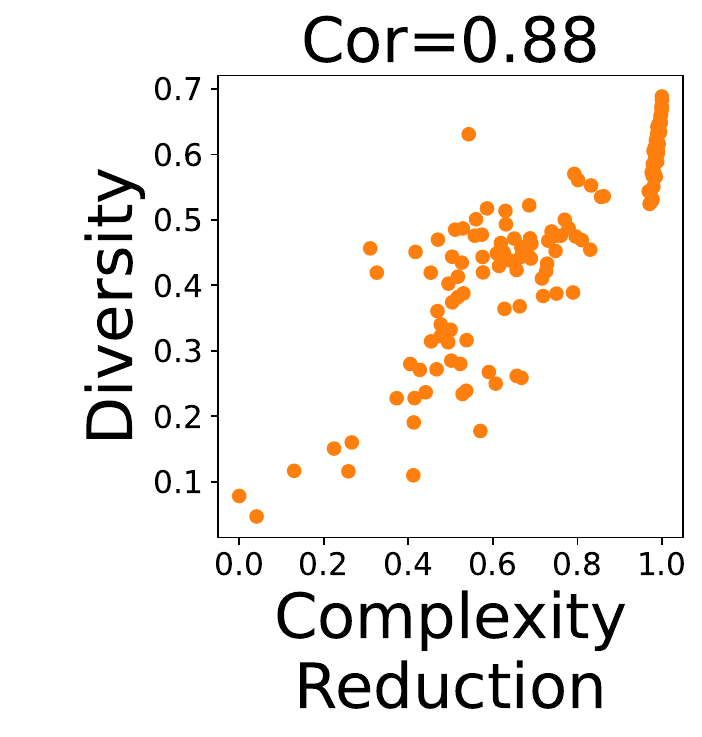}};
		\end{tikzpicture}
	\end{subfigure}
	\begin{subfigure}[b]{0.24\linewidth}
		\begin{tikzpicture}
		\node (img)  {\includegraphics[width=\linewidth]{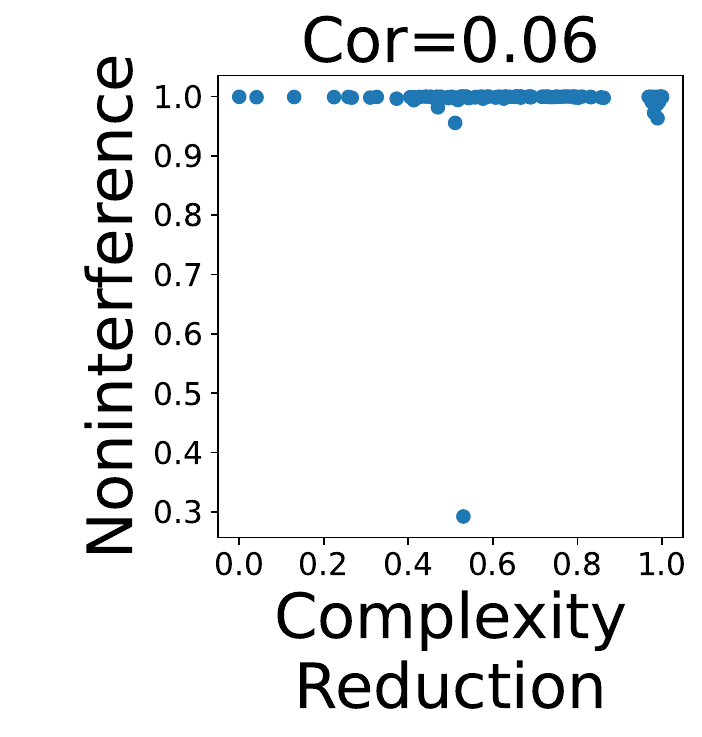}};
		\end{tikzpicture}
	\end{subfigure}
	\begin{subfigure}[b]{0.24\linewidth}
		\begin{tikzpicture}
		\node (img)  {\includegraphics[width=\linewidth]{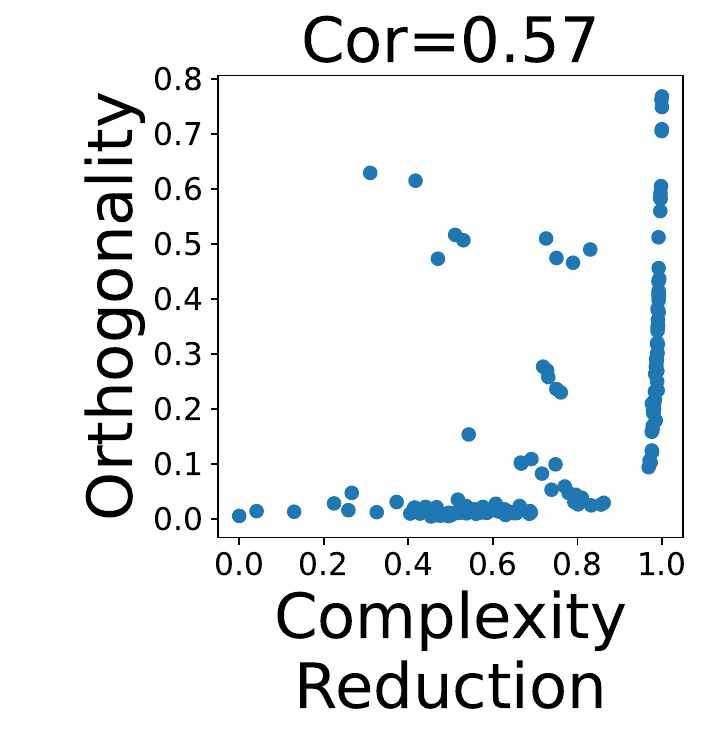}};
		\end{tikzpicture}
	\end{subfigure}
	\begin{subfigure}[b]{0.24\linewidth}
		\begin{tikzpicture}
		\node (img)  {\includegraphics[width=\linewidth]{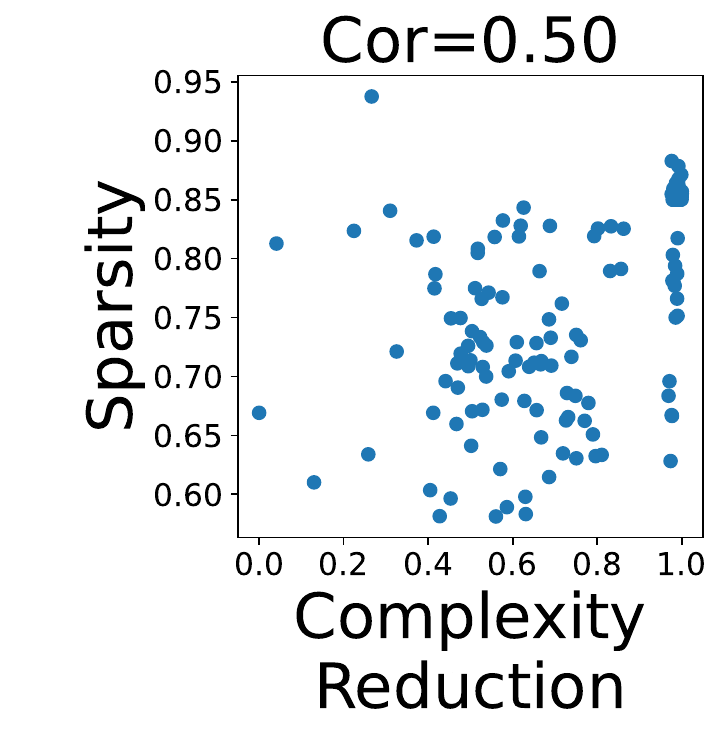}};
		\end{tikzpicture}
	\end{subfigure}
	
	\begin{subfigure}[b]{0.24\linewidth}
		\begin{tikzpicture}
		\node (img)  {\includegraphics[width=\linewidth]{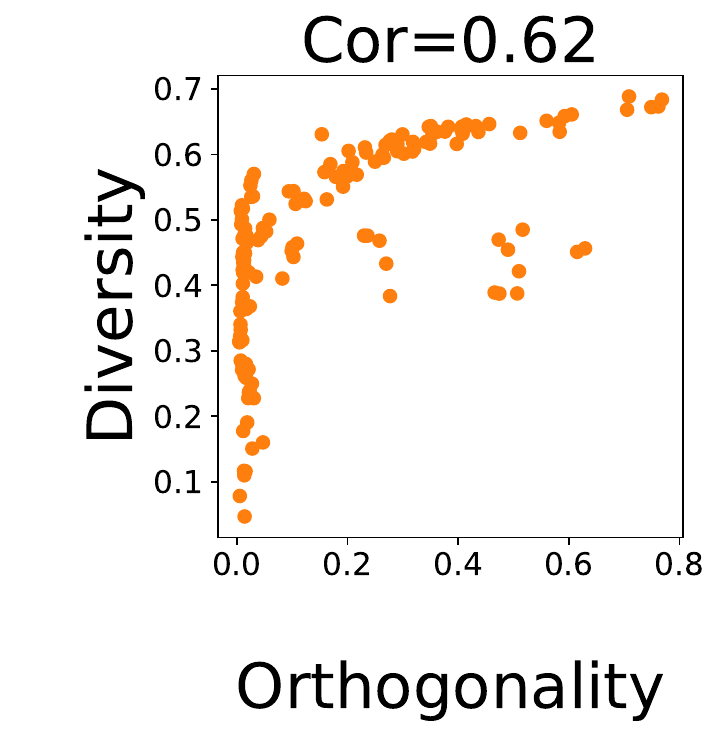}};
		\end{tikzpicture}
	\end{subfigure}
	\begin{subfigure}[b]{0.24\linewidth}
		\begin{tikzpicture}
		\node (img)  {\includegraphics[width=\linewidth]{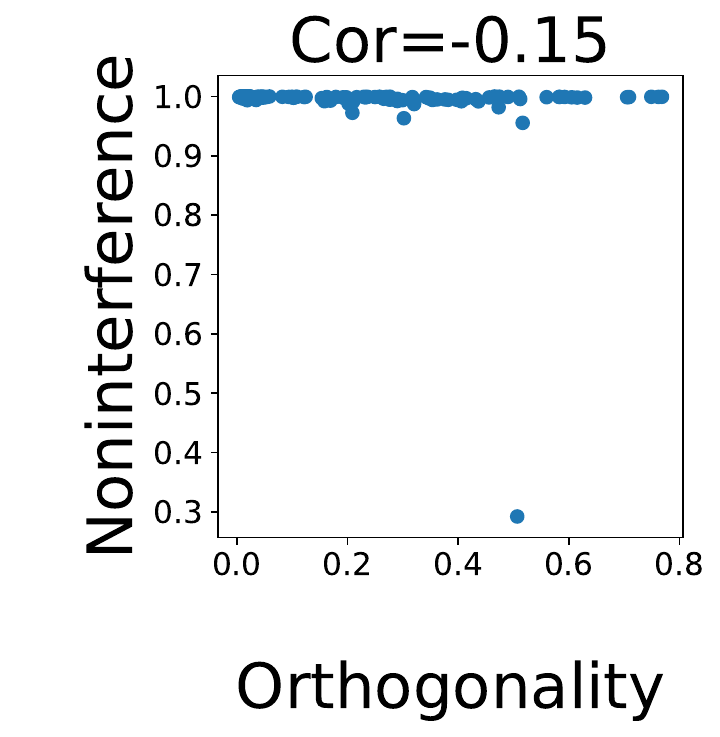}};
		\end{tikzpicture}
	\end{subfigure}
	\begin{subfigure}[b]{0.24\linewidth}
		\begin{tikzpicture}
		\node (img)  {\includegraphics[width=\linewidth]{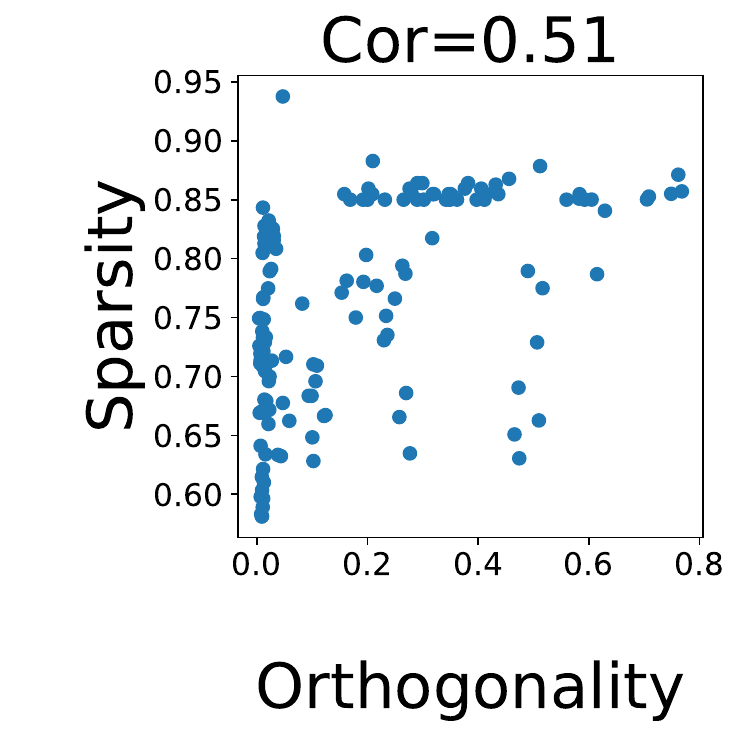}};
		\end{tikzpicture}
	\end{subfigure}

	\caption{There exists strong positive correlation between diversity and complexity reduction. Meanwhile, when diversity is high, the increment of diversity comes with an increment of orthogonality. The above 2 subplots are highlighted with orange color. Each subplot shows the relationship between a pair of properties. Each scatter refers to one representation. The x and y coordinates are property measures. The properties' names are shown as labels on each axis. The correlations (Cor) are reported above each subplt.}
	\label{fg:prop_pair}
\end{figure}
We also checked the relationship between properties. The result is shown in Figure \ref{fg:prop_pair}.
Two subplots are highlighted with the orange color. We noticed diversity and complexity reduction showed strong positive linear correlation. This suggests that monitoring either diversity or complexity reduction should be informative to predict the dissimilar task transfer performance in practice.
Furthermore, a threshold exists when looking at diversity and orthogonality, as well as diversity and complexity reduction. For representations with higher diversity (higher than 0.5, in this case), it also showed higher orthogonality, while this pattern does not exist in low diversity representations. Although there exists several outliers, this still indicates the possibility that pursuing a representation with high orthogonality may result in a relatively high diversity at the same time in practice.

\section{Experimental Details for Atari}\label{app_atari}

We used Rainbow-DQN with the default hyper-parameter setting provided by Hessel et al. \cite{hessel2018rainbow}.
The rainbow agent combines DQN \cite{mnih2015human} double Q-learning \cite{hasselt2016deep}, prioritized experience replay with the proportional variant \cite{schaul2015prioritized}, dueling networks \cite{wang2016dueling}, noisy networks for exploration \cite{fortunato2018noisy}, and distributional RL \cite{bellemare2017distributional}.
The agent is similar to the Rainbow-IQN agent used in the recent work showing transfer across modes \cite{rusu2022probing} except that it uses C51 for distributional RL \cite{bellemare2017distributional} instead of IQN \cite{dabney2018distributional}. 

We interpret the convolutional layers of the network of the Rainbow agent as the representation network and the successive fully connecting layers as the value network. The architectural details are worth mentioning here.
The representation network consists of three convolution layers each followed by ReLU activations, and produces a representation of 3136 features, as a function of stacked input frames (see Hessel et al. \cite{hessel2018rainbow} for the number of kernels and stride configuration).
The value-network consists of separate fully-connected networks that estimate value and advantage \cite{wang2016dueling}.
The value network takes as input the output of the representation network and the outputs of the value network are combined to estimate the value distribution \cite{bellemare2017distributional}.
The fully-connected networks that estimate value and advantage have a single hidden layer of 512 units, with noisy linear networks \cite{fortunato2018noisy} and ReLU activations.

\section{Additional Complexity Results in Atari}\label{app_complexity}
We put the raw complexity in Atari in Table \ref{tb:raw-capacity}
\begin{table}[!t]
    \renewcommand{\arraystretch}{1.3}
    \caption{Unnormalized complexity (L\_rep) of the learned and random Rainbow representations in Freeway, Space Invaders, and Breakout (average of 5 representations)}
    \label{tb:raw-capacity}
    \centering
    \begin{tabular}{c c c}
        Game & Learned Representation & Random Representation  \\ 
        \hline
        Freeway  & 0.029 & 4.178 \\
        Space Invaders & 0.228 & 6.908 \\
        Breakout  & 0.229 & 6.034 \\
    \end{tabular}
\end{table}


\end{document}